\pdfoutput=1
\documentclass{article}

     \usepackage[nonatbib, final]{tackling_climate_workshop_style}

\usepackage[utf8]{inputenc} %
\usepackage[T1]{fontenc}    %
\usepackage{hyperref}       %
\usepackage{url}            %
\usepackage{booktabs}       %
\usepackage{amsfonts}       %
\usepackage{nicefrac}       %
\usepackage{microtype}      %

\usepackage{graphicx}

\usepackage{xcolor}
\usepackage[normalem]{ulem}
\newif\ifshowchanges
\showchangestrue

\title{Evaluating the Effectiveness of Location Encoding Methods for Species Distribution Modelling}

\author{  \textbf{Christian Lange} \quad  \quad
  \textbf{Oisin Mac Aodha}\\
  \\
  University of Edinburgh
}

\begin{document}

\maketitle

\begin{abstract}
   Understanding the spatial distribution of species is a key question in ecology. A variety of solutions, from statistical, traditional machine learning, through to deep learning-based methods have been proposed. These methods take sparse observations for a species of interest as input and aim to generate dense spatial outputs indicating the spatial range of the species. Orthogonal to the choice of statistical model used is the question of what input features are most informative. Thanks to recent developments, the number of options has greatly expanded, with choices spanning different climate variables, learning-based encodings, earth embedding models, or raw transformed geographic coordinates. In this work, we introduce a standardized evaluation framework and quantify the effectiveness of  different location encodings in the context of species range estimation, both at the local and global scale.
\end{abstract}

\section{Introduction}
Understanding where different species of plants and animals are spatially distributed and how these distributions change over time is an important question in both ecological research and in conservation practice. 
The key challenge is that obtaining sufficient spatial coverage to say whether a species of interest is  present or absent at a specific location is extremely costly and time consuming. 
To address this, species distribution models (SDMs) provide a computational way to convert sparse observations into dense spatial predictions~\cite{beery2021species}. 

SDMs aim to learn a mapping from encodings of locations on earth to numerical quantities of interest (e.g., is a species present or not or how many individuals of that species are there at that location). 
Traditionally, the go-to representation/encoding of different locations are environmental/climatic variables, e.g., average rainfall, number of sunlight hours, elevation, land type, etc.  
However, if the input encoding is not sufficiently expressive it is not possible to learn an accurate mapping from inputs to outputs. 
As a result, this has a knock-on effect on our ability to make effective and informed conservation and land management decisions. 

In the last few years, we have seen the development of new approaches for geographic representation learning~\cite{klemmer2026earth}.  
Powered by the strengths of deep learning-based backbones and the wide availability of geographic data (e.g., aerial images, satellite derived data products such as land use, and on the ground data with location meta-data) these methods have been demonstrated to learn effective representations that can be used for a number of downstream tasks. 
In general terms, these methods can generate a feature encoding for a given query location that can be interpreted as a general-purpose representation of that location.  

In this work, we evaluate how effective different types of location encoding methods are for estimating the spatial range of different species.  
Specifically, we evaluate a suite of different families of encodings, spanning simple coordinate transformations to complex learning-based backbones, across both global and regional datasets. 
Our results highlight some interesting observations that are relevant for downstream practitioners that aim to use these techniques in different climate and conservation relevant use cases.  

\section{Method}
Our objective is to evaluate the effectiveness of different types of features for the task of species range prediction.
Here we outline our evaluation framework, including the features used, how they were probed, and the evaluation data used. 

\noindent{\textbf{Features}.} We compare several different families of features spanning (i) raw coordinates and their transformations (i.e., latitude and longitude coordinates, sin and cos transformations~\cite{mac2019presence}, and random Fourier projections~\cite{tancik2020fourier}), (ii) environmental features e.g., ERA5 and Bioclim), (iii) earth embedding models (e.g., SatCLIP~\cite{klemmer2025satclip},  GeoCLIP~\cite{vivanco2023geoclip}, AlphaEarth~\cite{brown2025alphaearth} and TESSERA~\cite{feng2026tessera}), and (iv) features derived from deep learning-based SDMs (e.g., SINR~\cite{cole2023spatial}). 
With the exception of the SINR models, which we retrain to remove the species from the evaluation sets, the other models are used as is (see Appendix~\ref{app:features}). 

\noindent{\textbf{Probing}.} Taking inspiration from evaluation protocols commonly used in the representation learning literature, we freeze the input features and use simple probes to evaluate how effective different features are. 
Specifically, we report performance using down-sampled Random Forest~\cite{valavi2021modelling}, Logistic Regression, and MaxEnt~\cite{phillips2004maximum} probes (Appendix~\ref{app:protocol}), where a separate probe is trained for each species. 
Given that the most commonly available form of biodiversity observation data is presence-only (i.e., only confirmed detections of an organism are recorded), we use pseudo-absences (i.e., randomly selected locations) as the training absences. 
We also report performance for a feature-free null baseline that scores a location by its distance to the nearest training presence~\cite{hijmans2012cross}.

\noindent{\textbf{Evaluation}.} We selected evaluation data that spans global, national, and state-level scales. 
For the global data, like SINR~\cite{cole2023spatial}, we select expert-derived range maps from the IUCN~\cite{IUCNRedListData}. 
Specifically, we select  3,052 species across 12 groups (mammals, birds, reptiles, amphibians, plants, and seven freshwater groups).  
Species were selected by requiring that 90\% of their iNaturalist~\cite{iNatWeb} observations fall within the expert range map. 
Ranges are rasterised onto H3~\cite{H3Web} cells of resolution 5 (e.g., $\sim~$253 km$^2$ each), and for each cell we denote it as either being a true presence or true absence for the species, depending on whether its centre lies inside the expert range map. 
While this data has limitations, see discussion in~\cite{cole2023spatial}, it represents the best available expert-derived global data. 
During training we use presence observations from iNaturalist~\cite{iNatWeb} for the corresponding species. 

For the state/national data, we selected 37 species of bats and butterflies from the UK and  54 amphibian and snake species from California.
Presence-only UK-based observation data was obtained from the NBN Atlas~\cite{nbnatlas} and California data is from iNaturalist (see Appendix~\ref{app:datasets}). 
Unlike the IUCN data, there exist no expert range maps for all of these species, and thus we opt to hold out some of the observations  to form a test set. 
While this is not ideal, it represents a pragmatic compromise, and given the highly monitored status of the UK and California there is sufficient density of observations to get sufficient coverage. 
The data is filtered to remove exact duplicate coordinates, keeping only one observation per H3 resolution eight cell, and we only retain observations that occur on land.  
Finally, we use five-fold spatially blocked cross-validation~\cite{roberts2017cross} over H3 resolution three cells, with test absences at locations where only other species of the same group were recorded~\cite{phillips2009sample}. 

\section{Results}

In Table~\ref{tab:main_res} we report quantitative results for global (3,052 species), UK (37 species), and California (54) species for the different families of features.  
It was not feasible for us to download AlphaEarth or TESSERA embeddings for all locations of earth, and thus they are omitted for the global evaluation. 
The results illustrate that the SDM and earth embedding-based features tend to be the best on average, but that simple transformations of the raw coordinates (e.g., `rand fourier'), are surprisingly effective. 
Depending on the metric used, the null baseline can outperform most feature sets globally and is competitive locally.
Appendix~\ref{app:results} provides additional metrics and discussion on these.

\begin{table}[t]
\vspace{-5pt}
  \caption{Quantitative evaluation of different location encodings using Random Forest (RF), Logistic Regression (LR), or MaxEnt (ME) probes on the task of species range estimation. Performance is reported using Average Precision (AP). Type: location (loc), environmental (env), earth embedding (ee), or SDM-derived (sdm). Bold: best feature set per column. $^{\dagger}$one to three species not fitted.}
  \label{tab:main_res}
\vspace{0pt}
\resizebox{\textwidth}{!}{
\begin{tabular}{lcr|rrl|rrl|rrl}
 & &  & \multicolumn{3}{|c|}{\textbf{Global}} & \multicolumn{3}{|c|}{\textbf{UK}} & \multicolumn{3}{|c}{\textbf{California}} \\
\textbf{Features} & \textbf{Type} & \textbf{Dim} & \textbf{RF} & \textbf{LR} & \textbf{ME} & \textbf{RF} & \textbf{LR} & \textbf{ME} & \textbf{RF} & \textbf{LR} & \textbf{ME} \\ \hline
coords & loc & 2 & 0.72 & 0.16 & 0.58$^{\dagger}$ & 0.69 & 0.63 & 0.66 & 0.56 & 0.47 & 0.54 \\
sin cos & loc & 16 & 0.72 & 0.66 & 0.69 & 0.68 & 0.65 & 0.67 & 0.55 & 0.52 & 0.55 \\
rand fourier & loc & 128 & 0.76 & 0.67 & \textbf{0.71} & 0.69 & 0.67 & 0.68 & 0.59 & 0.55 & 0.56 \\
ERA5 & env & 117 & 0.63 & 0.51 & 0.50 & 0.68 & 0.68 & 0.67 & 0.60 & 0.56 & 0.54 \\
bioclim & env & 20 & 0.54 & 0.24 & 0.47 & 0.66 & 0.65 & 0.66 & 0.61 & 0.58 & 0.61 \\
CHELSA & env & 19 & 0.50 & 0.22 & 0.44 & 0.66 & 0.64 & 0.66 & 0.58 & 0.53 & 0.58 \\
SoilGrids & env & 14 & 0.41 & 0.23 & 0.36$^{\dagger}$ & 0.67 & 0.63 & 0.65 & 0.54 & 0.46 & 0.53 \\
Human Footprint & env & 9 & 0.13 & 0.09 & 0.12$^{\dagger}$ & 0.61 & 0.59 & 0.61 & 0.39 & 0.37 & 0.38$^{\dagger}$ \\
MOSAIKS & ee & 512 & 0.57 & 0.39 & 0.37$^{\dagger}$ & 0.68 & 0.65 & 0.66 & 0.58 & 0.52 & 0.55 \\
AlphaEarth 1 km & ee & 64 & -- & -- & -- & 0.69 & 0.70 & 0.68 & 0.65 & 0.65 & 0.63 \\
AlphaEarth multi-scale & ee & 192 & -- & -- & -- & 0.69 & 0.70 & 0.68 & 0.67 & \textbf{0.67} & \textbf{0.65} \\
TESSERA 1 km & ee & 128 & -- & -- & -- & 0.68 & 0.69 & 0.67 & 0.61 & 0.62 & 0.59 \\
sin cos + AlphaEarth 1 km & loc + ee & 80 & -- & -- & -- & \textbf{0.71} & \textbf{0.71} & \textbf{0.70} & \textbf{0.68} & \textbf{0.67} & \textbf{0.65} \\
sin cos + TESSERA 1 km & loc + ee & 144 & -- & -- & -- & 0.70 & 0.70 & 0.67 & 0.66 & 0.63 & 0.59 \\
SatCLIP & ee & 256 & \textbf{0.77} & \textbf{0.68} & \textbf{0.71} & 0.70 & 0.67 & 0.68 & 0.60 & 0.55 & 0.57 \\
GeoCLIP & ee & 512 & 0.71 & 0.66 & 0.64 & 0.68 & 0.67 & 0.68 & 0.61 & 0.56 & 0.59 \\
SINR global & sdm & 256 & \textbf{0.77} & 0.61 & 0.64 & 0.69 & 0.66 & 0.67 & 0.63 & 0.58 & 0.59 \\
SINR UK / California & sdm & 256 & -- & -- & -- & 0.70 & 0.68 & 0.68 & 0.65 & 0.63 & 0.62 \\ \hline
null (dist. to nearest obs.) & baseline & -- & \multicolumn{3}{|c|}{0.75} & \multicolumn{3}{|c|}{0.66} & \multicolumn{3}{|c}{0.54} \\ 
\end{tabular}
}
\end{table}

\begin{figure}[t]
\centering
\includegraphics[width=1.0\textwidth]{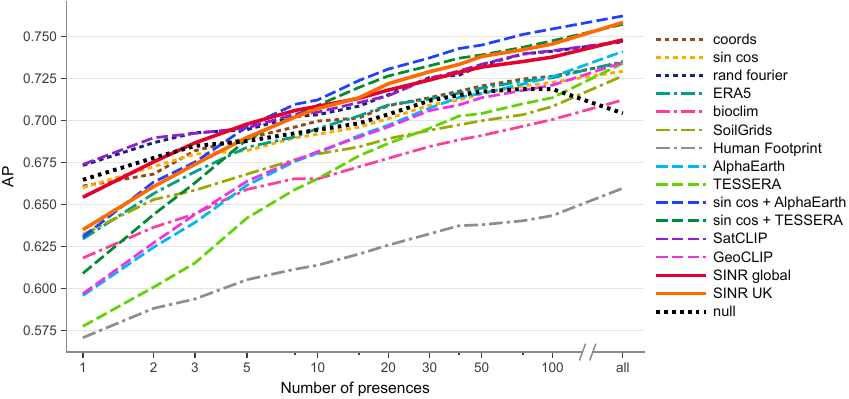}
\vspace{-10pt}
\caption{Few-shot evaluation for the UK data where we vary the amount of presence observations available during training. 
Average Precision (AP) results are reported using a Random Forest probe, averaged over 34 species. `all' uses every training presence available.}
\label{fig:uk_few_shot}
\end{figure}

\begin{figure}[h]
\centering
\includegraphics[width=1.0\textwidth]{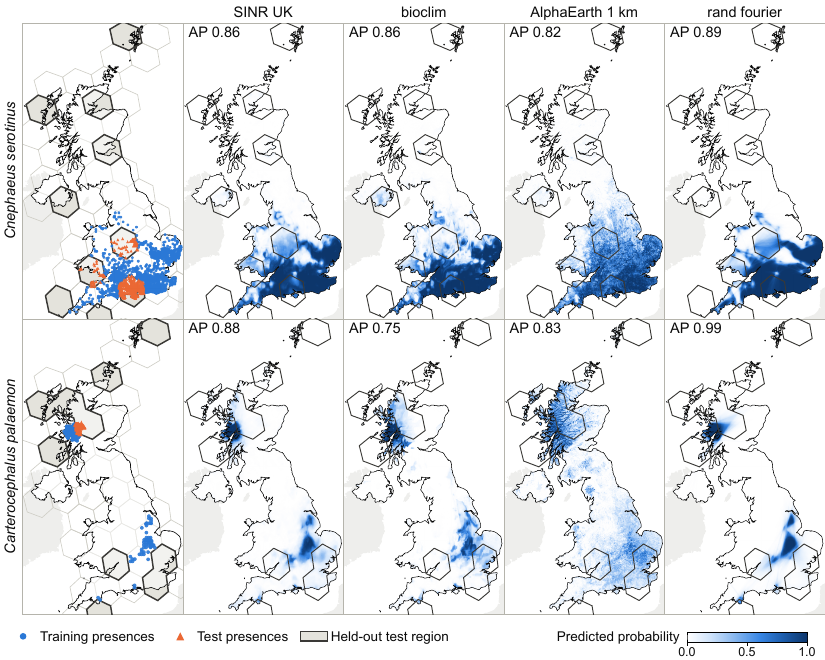}
\vspace{-10pt}
\caption{Predicted UK ranges for one bat (\emph{C.~serotinus}, top) and one butterfly (\emph{C.~palaemon}, bottom) species using a Random Forest probe. Left: training (blue) and test (orange) presences and the held-out region of the fold shown; the Average Precision (AP) reported is for that fold.}
\label{fig:uk_qual_1} %
\end{figure}

In Figure~\ref{fig:uk_few_shot} we explore a few-shot evaluation setting relevant to practitioners whereby we vary the number of presence observations during training for the UK dataset. 
We observe that the SDM-derived features (`SINR UK') are competitive with AlphaEarth features, and that the SINR model trained exclusively on UK data performs slightly better than a SINR model trained on global data once roughly 20 or more presences are available (see `SINR UK' versus `SINR global').
Simple coordinate-based features (coords, sin cos) are quite effective with very few presences, and the more expressive coordinate Fourier encoding is competitive at higher presence amounts in the absence of the availability of a sophisticated location encoding model.
Example predictions for a subset of input features are shown in Figure~\ref{fig:uk_qual_1}.

\section{Discussion}
This work does not present a new type of location encoding model. 
Instead it evaluates how effective different existing features are. 
We view this as a starting point for the development of future models that generate representations that are more effective for species distribution-centric tasks. 

Evaluating species distribution models is a challenging task owing to the lack of high quality globally distributed reference data. 
As a result, practitioners are forced to use proxies such as expert-derived maps or held out data as in this work. 
Future work would benefit from globally coordinated efforts to amass reliable data that spans a large number of different types of species. 

While we observe that earth embedding methods perform well overall, they are not all equal, e.g., AlphaEarth is often superior to SatCLIP/GeoCLIP. 
However, this could be partially attributed to some contamination in the training data as AlphaEarth uses GBIF species occurrence records in training, which could potentially include the evaluation species.  
There are also concerns related to the energy use associated with training large earth embedding models, and as a result the relatively efficient SDMs such as SINR, that do not make use of expensive satellite images, could be a better trade-off.

\section{Conclusion}
We presented a detailed global and local evaluation of the effectiveness of different types of location encoding methods for the task of species range estimation. 
We observed a general trend that more complex learning-based methods resulted in superior performance, especially regionally. 
However, simple transformations of raw coordinates were also effective, particularly with very few known presences. 
In future work, we aim to expand the evaluation both in terms of the number of location encoding methods, evaluation metrics, and the number of species.  

\newpage
{\small
\bibliographystyle{plain}
\bibliography{main}

\begin{thebibliography}{10}

\bibitem{c3s2021bioclimatic}
{Copernicus Climate Change Service. Global bioclimatic indicators from 1979 to
  2018 derived from reanalysis}.
\newblock Climate Data Store, \url{https://doi.org/10.24381/cds.bce175f0},
  2021.

\bibitem{H3Web}
{H3}.
\newblock \url{https://h3geo.org/}, accessed 1 Aug 2026.

\bibitem{iNatWeb}
{iNaturalist}.
\newblock \url{www.inaturalist.org}, accessed 1 Aug 2026.

\bibitem{IUCNRedListData}
{IUCN 2022. The IUCN Red List of Threatened Species. 2022-2}.
\newblock \url{https://www.iucnredlist.org}, accessed 1 Aug 2026.

\bibitem{nbnatlas}
{National Biodiversity Network (NBN Atlas)}.
\newblock \url{https://nbnatlas.org}, accessed 1 Aug 2026.

\bibitem{naturalearth}
{Natural Earth}.
\newblock \url{https://www.naturalearthdata.com}, accessed 15 Jun 2026.

\bibitem{beery2021species}
Sara Beery, Elijah Cole, Joseph Parker, Pietro Perona, and Kevin Winner.
\newblock Species distribution modeling for machine learning practitioners: A
  review.
\newblock In {\em Conference on Computing and Sustainable Societies}, 2021.

\bibitem{brown2025alphaearth}
Christopher~F Brown, Michal~R Kazmierski, Valerie~J Pasquarella, William~J
  Rucklidge, Masha Samsikova, Chenhui Zhang, Evan Shelhamer, Estefania Lahera,
  Olivia Wiles, Simon Ilyushchenko, et~al.
\newblock Alphaearth foundations: An embedding field model for accurate and
  efficient global mapping from sparse label data.
\newblock {\em arXiv:2507.22291}, 2025.

\bibitem{cole2023spatial}
Elijah Cole, Grant Van~Horn, Christian Lange, Alexander Shepard, Patrick Leary,
  Pietro Perona, Scott Loarie, and Oisin Mac~Aodha.
\newblock Spatial implicit neural representations for global-scale species
  mapping.
\newblock In {\em ICML}, 2023.

\bibitem{elithreview}
Jane Elith and John~R. Leathwick.
\newblock Species distribution models: Ecological explanation and prediction
  across space and time.
\newblock {\em Annual Review of Ecology, Evolution, and Systematics}, 2009.

\bibitem{feng2026tessera}
Zhengpeng Feng, Clement Atzberger, Sadiq Jaffer, Jovana Knezevic, Silja
  Sormunen, Robin Young, Madeline~C Lisaius, Markus Immitzer, Toby Jackson,
  James Ball, et~al.
\newblock Tessera: Temporal embeddings of surface spectra for earth
  representation and analysis.
\newblock In {\em CVPR}, 2026.

\bibitem{fick2017worldclim}
Stephen~E Fick and Robert~J Hijmans.
\newblock {WorldClim} 2: new 1-km spatial resolution climate surfaces for
  global land areas.
\newblock {\em International Journal of Climatology}, 2017.

\bibitem{hijmans2012cross}
Robert~J Hijmans.
\newblock Cross-validation of species distribution models: removing spatial
  sorting bias and calibration with a null model.
\newblock {\em Ecology}, 2012.

\bibitem{karger2017climatologies}
Dirk~Nikolaus Karger, Olaf Conrad, J{\"u}rgen B{\"o}hner, Tobias Kawohl, Holger
  Kreft, Rodrigo~Wilber Soria-Auza, Niklaus~E Zimmermann, H~Peter Linder, and
  Michael Kessler.
\newblock Climatologies at high resolution for the earth's land surface areas.
\newblock {\em Scientific Data}, 2017.

\bibitem{klemmer2025satclip}
Konstantin Klemmer, Esther Rolf, Caleb Robinson, Lester Mackey, and Marc
  Ru{\ss}wurm.
\newblock {SatCLIP}: Global, general-purpose location embeddings with satellite
  imagery.
\newblock In {\em AAAI}, 2025.

\bibitem{klemmer2026earth}
Konstantin Klemmer, Esther Rolf, Marc Russwurm, Gustau Camps-Valls, Mikolaj
  Czerkawski, Stefano Ermon, Alistair Francis, Nathan Jacobs, Hannah Kerner,
  Lester Mackey, et~al.
\newblock Earth embeddings: Toward artificial intelligence-centric
  representations of our planet.
\newblock {\em Geoscience and Remote Sensing Magazine}, 2026.

\bibitem{kornblith2019similarity}
Simon Kornblith, Mohammad Norouzi, Honglak Lee, and Geoffrey Hinton.
\newblock Similarity of neural network representations revisited.
\newblock In {\em ICML}, 2019.

\bibitem{mac2019presence}
Oisin Mac~Aodha, Elijah Cole, and Pietro Perona.
\newblock Presence-only geographical priors for fine-grained image
  classification.
\newblock In {\em ICCV}, 2019.

\bibitem{phillips2017opening}
Steven~J Phillips, Robert~P Anderson, Miroslav Dud{\'\i}k, Robert~E Schapire,
  and Mary~E Blair.
\newblock Opening the black box: an open-source release of {Maxent}.
\newblock {\em Ecography}, 2017.

\bibitem{phillips2009sample}
Steven~J Phillips, Miroslav Dud{\'\i}k, Jane Elith, Catherine~H Graham, Anthony
  Lehmann, John Leathwick, and Simon Ferrier.
\newblock Sample selection bias and presence-only distribution models:
  implications for background and pseudo-absence data.
\newblock {\em Ecological Applications}, 2009.

\bibitem{phillips2004maximum}
Steven~J Phillips, Miroslav Dud{\'\i}k, and Robert~E Schapire.
\newblock A maximum entropy approach to species distribution modeling.
\newblock In {\em ICML}, 2004.

\bibitem{poggio2021soilgrids}
Laura Poggio, Luis~M de~Sousa, Niels~H Batjes, Gerard B~M Heuvelink, Bas
  Kempen, Eloi Ribeiro, and David Rossiter.
\newblock {SoilGrids} 2.0: producing soil information for the globe with
  quantified spatial uncertainty.
\newblock {\em SOIL}, 2021.

\bibitem{roberts2017cross}
David~R Roberts, Volker Bahn, Simone Ciuti, Mark~S Boyce, Jane Elith, Gurutzeta
  Guillera-Arroita, Severin Hauenstein, Jos{\'e}~J Lahoz-Monfort, Boris
  Schr{\"o}der, Wilfried Thuiller, et~al.
\newblock Cross-validation strategies for data with temporal, spatial,
  hierarchical, or phylogenetic structure.
\newblock {\em Ecography}, 2017.

\bibitem{rolf2021generalizable}
Esther Rolf, Jonathan Proctor, Tamma Carleton, Ian Bolliger, Vaishaal Shankar,
  Miyabi Ishihara, Benjamin Recht, and Solomon Hsiang.
\newblock A generalizable and accessible approach to machine learning with
  global satellite imagery.
\newblock {\em Nature Communications}, 2021.

\bibitem{tancik2020fourier}
Matthew Tancik, Pratul Srinivasan, Ben Mildenhall, Sara Fridovich-Keil, Nithin
  Raghavan, Utkarsh Singhal, Ravi Ramamoorthi, Jonathan Barron, and Ren Ng.
\newblock Fourier features let networks learn high frequency functions in low
  dimensional domains.
\newblock {\em NeurIPS}, 2020.

\bibitem{valavi2021modelling}
Roozbeh Valavi, Jane Elith, Jos{\'e}~J Lahoz-Monfort, and Gurutzeta
  Guillera-Arroita.
\newblock Modelling species presence-only data with random forests.
\newblock {\em Ecography}, 2021.

\bibitem{venter2016global}
Oscar Venter, Eric~W Sanderson, Ainhoa Magrach, James~R Allan, Jutta Beher,
  Kendall~R Jones, Hugh~P Possingham, William~F Laurance, Peter Wood,
  Bal{\'a}zs~M Fekete, et~al.
\newblock Global terrestrial {Human Footprint} maps for 1993 and 2009.
\newblock {\em Scientific Data}, 2016.

\bibitem{vivanco2023geoclip}
Vicente Vivanco~Cepeda, Gaurav~Kumar Nayak, and Mubarak Shah.
\newblock {GeoCLIP}: {CLIP}-inspired alignment between locations and images for
  effective worldwide geo-localization.
\newblock {\em NeurIPS}, 2023.

\end{thebibliography}
}

\newpage
\appendix
\setcounter{table}{0}   
\renewcommand{\thetable}{A\arabic{table}}
\setcounter{figure}{0}
\renewcommand{\thefigure}{A\arabic{figure}}
\renewcommand{\theHtable}{A\arabic{table}}\renewcommand{\theHfigure}{A\arabic{figure}} %
\section{Datasets and methodology}
\label{app:methods}

\subsection{Datasets}
\label{app:datasets}

Each dataset consists of presence observations for a set of species, which are filtered as described below. How absences are generated and how the data are split for cross-validation is described in Section~\ref{app:protocol}.

\paragraph{UK bats and butterflies.}
Presence observations are downloaded from the NBN Atlas~\cite{nbnatlas} for 37 species, 17 bats, and 20 butterflies (Table~\ref{tab:uk_species}).
This data is filtered to include only one record per species per H3 resolution eight cell (about 0.7~km$^2$) inside the United Kingdom polygon of Natural Earth~\cite{naturalearth}, which excludes sea, Ireland, and the Channel Islands.
No filter on date or on stated coordinate uncertainty is applied. This leaves 323{,}487 presences. 
The number of presences per species is very uneven: the minimum is 28 (\textit{Myotis alcathoe}), the median 2{,}791, and the maximum 44{,}595 (\textit{Aphantopus hyperantus}); bats and butterflies have a median of 1{,}231 and  of 5{,}928 presences respectively.
The observations were collected over more than a century (1900--2026; median year 2010): 28\% date from before 2000 and 65\% from before 2015.
A species' presences can therefore include places where it no longer occurs, so the dataset describes a historical rather than the current distribution of each species.
Few-shot results (varying the number of training presences up to 100) leave out three species whose training folds contain too few presences to reach 100: \textit{Myotis alcathoe} (28 presences), \textit{Phengaris arion} (37) and \textit{Plecotus austriacus} (62).

\begin{table}[h]
\centering\scriptsize
\caption{UK species and their number of presences $n$ after filtering (323{,}487 in total). B = bat, Bf = butterfly. Sorted by group, then by $n$.}\label{tab:uk_species}
\setlength{\tabcolsep}{3pt}
\resizebox{\textwidth}{!}{\begin{tabular}{l l r@{\hspace{6pt}}l l r@{\hspace{6pt}}l l r}
\hline
Species & Group & $n$ & Species & Group & $n$ & Species & Group & $n$ \\
\hline
\textit{Pipistrellus pygmaeus} & B & 10{,}894 & \textit{Pipistrellus pipistrellus} & B & 8{,}778 & \textit{Plecotus auritus} & B & 8{,}284 \\
\textit{Myotis daubentonii} & B & 6{,}047 & \textit{Nyctalus noctula} & B & 5{,}631 & \textit{Myotis nattereri} & B & 3{,}216 \\
\textit{Rhinolophus hipposideros} & B & 2{,}791 & \textit{Cnephaeus serotinus} & B & 2{,}304 & \textit{Rhinolophus ferrumequinum} & B & 1{,}231 \\
\textit{Myotis mystacinus} & B & 1{,}191 & \textit{Nyctalus leisleri} & B & 991 & \textit{Barbastella barbastellus} & B & 856 \\
\textit{Pipistrellus nathusii} & B & 700 & \textit{Myotis brandtii} & B & 346 & \textit{Myotis bechsteinii} & B & 180 \\
\textit{Plecotus austriacus} & B & 62 & \textit{Myotis alcathoe} & B & 28 & & & \\ \hline \textit{Aphantopus hyperantus} & Bf & 44{,}595 &
\textit{Vanessa cardui} & Bf & 44{,}248  & \textit{Polyommatus icarus} & Bf & 40{,}202 \\ \textit{Thymelicus sylvestris} & Bf & 27{,}830 &
\textit{Celastrina argiolus} & Bf & 27{,}617 & \textit{Gonepteryx rhamni} & Bf & 26{,}364 \\ \textit{Melanargia galathea} & Bf & 14{,}034 &
\textit{Thymelicus lineola} & Bf & 12{,}406 & \textit{Argynnis paphia} & Bf & 9{,}957 \\ \textit{Satyrium w-album} & Bf & 7{,}338 &
\textit{Limenitis camilla} & Bf & 4{,}517 & \textit{Boloria euphrosyne} & Bf & 3{,}326 \\ \textit{Polyommatus coridon} & Bf & 2{,}205 &
\textit{Apatura iris} & Bf & 1{,}806 & \textit{Polyommatus bellargus} & Bf & 1{,}198 \\ \textit{Leptidea sinapis} & Bf & 933 &
\textit{Carterocephalus palaemon} & Bf & 730 & \textit{Leptidea juvernica} & Bf & 377 \\ \textit{Satyrium pruni} & Bf & 237 &
\textit{Phengaris arion} & Bf & 37 &  & & \\
\hline
\end{tabular}}
\end{table}

\paragraph{California amphibians and snakes.}
The data are research-grade iNaturalist~\cite{iNatWeb} observations of amphibians and snakes dated 1 January 2021 to 1 January 2026. 
The same filtering as for the UK is applied, and additionally observations whose stated positional accuracy is worse than 1~km are removed.
The species were manually selected to produce a diverse set and were only retained in the dataset if at least 50 presences remain after filtering.
The data contains 54 species, 23 amphibians and 31 snakes (Table~\ref{tab:ca_species}), and 49{,}263 presence observations in total.
The number of presences per species ranges from 53 (\textit{Crotalus atrox}) to 9{,}380 (\textit{Pseudacris regilla}), with a median of 423 (amphibians: 470, snakes: 383). In contrast to the UK data, all observations are recent (2021--2026), so the benchmark describes the current distribution as seen by iNaturalist observers.
The few-shot figure for California (Figure~\ref{fig:ca_few_shot}) includes the 40 of 54 species whose training folds reach 100 presences.

\begin{table}[h]
\centering\scriptsize
\caption{California species and their number of presences $n$ after filtering (49,263 in total). A = amphibian, S = snake. Sorted by group, then by $n$.}\label{tab:ca_species}
\setlength{\tabcolsep}{3pt}
\resizebox{\textwidth}{!}{\begin{tabular}{l l r@{\hspace{6pt}}l l r@{\hspace{6pt}}l l r}
\hline
Species & Group & $n$ & Species & Group & $n$ & Species & Group & $n$ \\
\hline
\textit{Pseudacris regilla} & A & 9{,}380 & \textit{Anaxyrus boreas} & A & 4{,}649 & \textit{Batrachoseps attenuatus} & A & 3{,}248 \\
\textit{Lithobates catesbeianus} & A & 1{,}917 & \textit{Aneides lugubris} & A & 1{,}908 & \textit{Taricha torosa} & A & 1{,}766 \\
\textit{Ensatina eschscholtzii} & A & 1{,}688 & \textit{Taricha granulosa} & A & 1{,}072 & \textit{Pseudacris cadaverina} & A & 681 \\
\textit{Batrachoseps major} & A & 585 & \textit{Batrachoseps nigriventris} & A & 500 & \textit{Taricha sierrae} & A & 470 \\
\textit{Dicamptodon ensatus} & A & 458 & \textit{Dicamptodon tenebrosus} & A & 419 & \textit{Aneides vagrans} & A & 174 \\
\textit{Aneides flavipunctatus} & A & 173 & \textit{Ambystoma gracile} & A & 145 & \textit{Anaxyrus punctatus} & A & 128 \\
\textit{Batrachoseps gavilanensis} & A & 128 & \textit{Xenopus laevis} & A & 102 & \textit{Spea intermontana} & A & 79 \\
\textit{Batrachoseps gregarius} & A & 62 & \textit{Ambystoma macrodactylum} & A & 58 & & & \\ \hline \textit{Pituophis catenifer} & S & 5{,}254 &
\textit{Lampropeltis californiae} & S & 3{,}224 & \textit{Contia tenuis} & S & 1{,}080 \\ \textit{Diadophis punctatus} & S & 1{,}017 &
\textit{Crotalus ruber} & S & 869 & \textit{Masticophis lateralis} & S & 852 \\ \textit{Crotalus oreganus} & S & 742 &
\textit{Charina bottae} & S & 630 & \textit{Thamnophis elegans} & S & 595 \\ \textit{Thamnophis atratus} & S & 460 &
\textit{Lichanura orcutti} & S & 450 & \textit{Thamnophis couchii} & S & 448 \\ \textit{Masticophis flagellum} & S & 435 &
\textit{Crotalus pyrrhus} & S & 427 & \textit{Thamnophis sirtalis} & S & 404 \\ \textit{Crotalus cerastes} & S & 383 &
\textit{Arizona occidentalis} & S & 274 & \textit{Crotalus scutulatus} & S & 227 \\ \textit{Rhinocheilus lecontei} & S & 192 &
\textit{Hypsiglena ochrorhynchus} & S & 183 & \textit{Coluber constrictor} & S & 179 \\ \textit{Indotyphlops braminus} & S & 168 &
\textit{Salvadora hexalepis} & S & 162 & \textit{Crotalus stephensi} & S & 127 \\ \textit{Phyllorhynchus decurtatus} & S & 117 &
\textit{Contia longicauda} & S & 116 & \textit{Sonora occipitalis} & S & 112 \\ \textit{Tantilla planiceps} & S & 103 &
\textit{Trimorphodon lyrophanes} & S & 98 & \textit{Rena humilis} & S & 92 \\ \textit{Crotalus atrox} & S & 53 & & & & & & \\
\hline
\end{tabular}}
\end{table}

\paragraph{Global species (IUCN range maps).}
The global dataset uses expert range maps from the IUCN Red List~\cite{IUCNRedListData} across 12 groups (birds, mammals, reptiles, amphibians, plants, and seven freshwater groups).
Only species for which at least 90\% of their iNaturalist observations fall inside the range polygon are used.
The IUCN range polygons are rasterised onto H3~\cite{H3Web} resolution five cells (about 253~km$^2$ each), where a cell is counted as a presence if its centre lies inside the polygon, and every other cell is an absence.
Probes are trained on iNaturalist observations, which are independent of the range maps.
We use the research-grade observations of the SINR training set~\cite{cole2023spatial} where the positional accuracy is 5~km or better, and keep one presence per species per occupied cell.
Training and testing occurs over the 525{,}010 H3 resolution five cells (26\% of the 2{,}016{,}842 cells that cover the globe) that have valid values for every feature set, of which 98\% are over land.
A species is only included in the dataset if it has at least five training cells and at least one range cell among the valid cells; this leaves 3{,}052 species in the dataset of the 3{,}159 species originally selected (Table~\ref{tab:global_groups}).

\begin{table}[h]
\centering\small
\caption{Globally distributed species per IUCN group, with the minimum / median / maximum number of training cells and of expert-range cells (out of 525{,}010 scored cells).}\label{tab:global_groups}
\begin{tabular}{l r r r}
\hline
Group & Species & Training cells & Range cells \\
\hline
Birds & 1{,}614 & 5 / 98 / 11{,}774 & 4 / 13{,}271 / 524{,}718 \\
Reptiles & 444 & 5 / 60 / 4{,}727 & 1 / 3{,}139 / 70{,}124 \\
Plants & 286 & 5 / 41 / 4{,}482 & 3 / 436 / 133{,}698 \\
Amphibians & 254 & 5 / 62 / 5{,}069 & 2 / 2{,}736 / 45{,}962 \\
Mammals & 228 & 8 / 91 / 8{,}080 & 16 / 9{,}249 / 138{,}575 \\
Dragonflies and damselflies & 106 & 5 / 83 / 2{,}295 & 19 / 19{,}615 / 313{,}016 \\
Freshwater fishes & 55 & 5 / 58 / 2{,}996 & 140 / 3{,}212 / 127{,}229 \\
Freshwater plants & 34 & 6 / 97 / 5{,}106 & 253 / 19{,}092 / 239{,}190 \\
Freshwater molluscs & 19 & 12 / 64 / 536 & 1{,}005 / 7{,}251 / 54{,}078 \\
Crayfish & 8 & 15 / 44 / 167 & 248 / 1{,}802 / 15{,}601 \\
Freshwater crabs & 3 & 16 / 36 / 42 & 82 / 408 / 1{,}338 \\
Freshwater shrimps & 1 & 22 / 22 / 22 & 7{,}130 / 7{,}130 / 7{,}130 \\
\hline
All groups & 3{,}052 & 5 / 77 / 11{,}774 & 1 / 6{,}899 / 524{,}718 \\
\hline
\end{tabular}
\end{table}

\subsection{Location features}
\label{app:features}

Table~\ref{tab:features} lists the 18 feature sets we compare.
Models used as feature extractors such as SINR and SatCLIP are frozen in all our experiments.
Environmental and satellite-derived features are stored as gridded rasters and read at each point by bilinear interpolation, only interpolating over cells with valid data and renormalizing the remaining weights.
Each continuous raster channel is z-scored once, over the valid cells of its own raster. The AlphaEarth and TESSERA embeddings are used as published.

\begin{table}[ht]
\centering
\caption{Feature sets. Type: loc = location encoding, env = environmental raster, ee = Earth-observation or pretrained embedding, sdm = species distribution model. Dim.\ is the number of features. ``Native'' is the resolution of the source data and ``Used'' the grid we sample. 1 arcmin is about 1.85\,km.}
\label{tab:features}
\resizebox{\textwidth}{!}{%
\begin{tabular}{llrlll}
\toprule
Name & Type & Dim. & Description & Native & Used \\
\midrule
coords & loc & 2 & longitude and latitude in degrees & -- & exact \\
sin cos & loc & 16 & sin and cos of longitude and latitude at 4 frequencies & -- & exact \\
rand fourier & loc & 128 & 64 random Fourier features of the 3-D position on the unit sphere & -- & exact \\
ERA5 & env & 117 & ERA5 biodiversity climatology (Copernicus CDS), 1979--2018 & 0.5$^\circ$ & 0.5$^\circ$ \\
bioclim & env & 20 & WorldClim 2.1 BIO1--19 and elevation, 1970--2000 & 5 arcmin & 5 arcmin \\
CHELSA & env & 19 & CHELSA v2.1 BIO1--19, 1981--2010 & 30 arcsec & 2.5 arcmin \\
SoilGrids & env & 14 & SoilGrids 2.0 soil properties, 0--5\,cm depth (carbon stock 0--30\,cm) & 250 m & 5 arcmin \\
Human Footprint & env & 9 & Human Footprint v2: index and 8 pressure layers, 2009 & 1 km & 5 arcmin \\
MOSAIKS & ee & 512 & random convolutional features of Planet imagery (2019 Q3), PCA from 4,000 & 0.25$^\circ$ & 0.25$^\circ$ \\
AlphaEarth 1 km & ee & 64 & AlphaEarth Foundations satellite embedding V1, 2025 & 10 m & 30 arcsec \\
AlphaEarth multi-scale & ee & 192 & AlphaEarth at 10, 30 and 150 arcsec, concatenated & 10 m & 10 arcsec--2.5 arcmin \\
TESSERA 1 km & ee & 128 & TESSERA v1 Sentinel-1/2 embedding, 2025 & 10 m & 30 arcsec \\
sin cos + AlphaEarth 1 km & loc + ee & 80 & concatenation of the two sets & -- & as above \\
sin cos + TESSERA 1 km & loc + ee & 144 & concatenation of the two sets & -- & as above \\
SatCLIP & ee & 256 & SatCLIP location encoder, ResNet50 image backbone, $L{=}10$ & -- & exact \\
GeoCLIP & ee & 512 & GeoCLIP location encoder, \texttt{geoclip} 1.2.0 & -- & exact \\
SINR global & sdm & 256 & penultimate layer of a SINR model trained worldwide & -- & exact \\
SINR UK / California & sdm & 256 & as above, trained only inside the region's bounding box & -- & exact \\
\bottomrule
\end{tabular}}
\end{table}

\paragraph{Coordinate encodings.}
\emph{coords} is longitude and latitude in degrees. \emph{sin cos} scales them to $x=\lambda/180$, $y=\varphi/90$ and returns $\sin$ and $\cos$ of $\pi 2^{l}x$ and $\pi 2^{l}y$ for $l=0,\dots,3$ (16 values), as in SINR~\cite{cole2023spatial}. \emph{rand fourier} maps the point to $p\in\mathbb{R}^3$ on the unit sphere and returns $\sin(2\pi B^{\top}p)$ and $\cos(2\pi B^{\top}p)$~\cite{tancik2020fourier}; $B\in\mathbb{R}^{3\times 64}$ has independent Gaussian entries with standard deviation 1.5, is drawn once from a fixed seed, and is the same for every species and fold.

\paragraph{Environmental rasters.}
\emph{bioclim} is the 19 WorldClim 2.1~\cite{fick2017worldclim} bioclimatic variables (BIO1--19) and elevation. \emph{CHELSA} is the 19 CHELSA v2.1~\cite{karger2017climatologies} bioclimatic variables, aggregated from 30 arcsec to 2.5 arcmin. \emph{ERA5} has 117 channels from the Copernicus ERA5 biodiversity indicators~\cite{c3s2021bioclimatic}: 19 bioclimatic variables; 13 temperature, precipitation, humidity and cloud indicators; 11 aridity and dry-spell; 10 evaporation; 10 surface heat-flux; 4 frost and growing-season; 5 soil-water; 5 wind; 10 sea-surface-temperature and sea-ice; and a 30-class one-hot K\"oppen--Geiger zone. Most are given for the annual mean and the warmest, coldest, wettest and driest quarters.
\emph{SoilGrids}~\cite{poggio2021soilgrids} has 13 properties of the 0--5\,cm layer (bulk density, cation exchange capacity, coarse fragments, clay, sand, silt, nitrogen, organic carbon density, soil organic carbon, pH, and water content at three suctions) and the 0--30\,cm organic carbon stock. \emph{Human Footprint}~\cite{venter2016global} is the composite 2009 index and its 8 pressure layers (built environment, cropland, pasture, night-time lights, navigable waterways, population density, railways, roads).

\paragraph{Satellite and learned embeddings.}
\emph{MOSAIKS}~\cite{rolf2021generalizable} reduces 4,000 random convolutional features per 0.25$^\circ$ cell to 512 principal components.
\emph{AlphaEarth}~\cite{brown2025alphaearth} is the annual 64-dimensional embedding for 2025; we average the unit-length vectors in each cell and renormalise, at 10 arcsec (0.3\,km), 30 arcsec (``1 km'') and 2.5 arcmin (``5 km''), and the multi-scale feature concatenates the three.
\emph{TESSERA}~\cite{feng2026tessera} is the 128-dimensional v1 embedding for 2025, averaged from 10\,m to 30 arcsec.
TESSERA and AlphaEarth exist only for the UK and California in our setup due to resource limitations.
\emph{SatCLIP}~\cite{klemmer2025satclip} is the location encoder of the ResNet50 model with spherical-harmonic degree $L=10$ (SIREN network, 256 outputs; input longitude, latitude). \emph{GeoCLIP}~\cite{vivanco2023geoclip} is the location encoder of the \texttt{geoclip} package, version 1.2.0 (512 outputs; input latitude, longitude).
Their image encoders are not used.

\paragraph{SINR models.}
The feature is the 256-dimensional last hidden layer of a SINR network~\cite{cole2023spatial}: a linear layer with ReLU and four residual blocks (two linear layers, ReLU, dropout 0.5), then a linear layer with one output per species.
The input is sin cos (16) concatenated with bioclim (20).
Training data are iNaturalist research-grade observations (April 2026 open-data export; 135.1 million observations of taxa with at least 50 observations), dropping those with accuracy worse than 5\,km and keeping one record per species and H3 resolution five cell.
The  AN\_FULL loss from ~\cite{cole2023spatial} is used and we use the same hyperparameters for training. 
The evaluation species from each dataset are not included in the SINR training data, so each test species has never been seen before by the SINR backbone.
 \emph{SINR UK} and \emph{SINR California} are trained only on observations and pseudo-absences inside each region's bounding box.
 Compared to ~\cite{cole2023spatial} coordinate inputs are renormalized to within the bounding box rather than over the whole globe, and the ``wrap'' is removed which connects the left and right hand side of the map, as this is only appropriate when modelling globally.

\paragraph{Concatenations.}
\emph{sin cos + AlphaEarth 1 km} and \emph{sin cos + TESSERA 1 km} append the embedding to the 16 sin and cos values.

\subsection{Probes, pseudo-absences and evaluation protocol}
\label{app:protocol}

\paragraph{Probes.}
One probe is fit for each fold for each species and feature set, with the same settings on all three datasets.

\emph{RF} is a down-sampled random forest \cite{valavi2021modelling}: 500 trees are used, each splitting on $\sqrt{d}$ random features (where $d$ is the input dimension) and trained on a bootstrap sample of $k$ presences plus a bootstrap sample of $k$ pseudo-absences, where $k$ is the number of training presences. The score is the mean predicted probability of presence.

\emph{LR} is an L2-regularized logistic regression ($C=1$, intercept, no class weights, L-BFGS, at most 1000 iterations) on features standardized with the training mean and standard deviation. 
The score is the predicted probability.

\emph{MaxEnt} is R \texttt{maxnet} 0.1.4~\cite{phillips2017opening} with default settings: regularization multiplier 1; linear features below 10 presences, adding quadratic below 15, hinge below 80 and pairwise products from 80; 50 hinge knots; pseudo-absence weight 100; presences also added to the background; predictions clamped to the training range; cloglog output.
Inputs with more than 20 dimensions are first reduced to 16 principal components fit on the training data.

The \emph{null} model~\cite{hijmans2012cross} uses no features and requires no training: it scores a location by the negative of the great-circle distance to the nearest training presence, so that points close to training presences have a higher score than those further away.

\paragraph{Pseudo-absences in the UK and California.}
Training pseudo-absences are drawn uniformly inside the region boundary.
For $n$ training presences, we draw $\mathrm{clip}(5n,1000,10000)$ pseudo-absences, i.e. we select five times $n$ random location inside the region, adding or removing locations so that we are left with between 1{,}000 and 10{,}000 pseudo-absences.
We also discard training pseudo-absences sharing an H3 resolution five cell with a training presence.
Test absences are the centres of H3 resolution six cells (about 36~km$^2$) inside the test area boundary that contain at least two observations of \emph{other} species of the same group and no observations of the evaluated species.
This process ensures that test absences are not selected at the same locations as actual test presence locations for the target species, and gives us a slightly increased likelihood that the target species is truly not present at the absence location.

\paragraph{Spatially blocked cross-validation.}
For our regional evaluation we perform five-fold spatial block cross validation.
Blocks are H3 resolution three cells (about 12{,}400~km$^2$).
Blocks with presences for the target species are shuffled, sorted by decreasing number of presences, and assigned in turn to the fold holding the fewest presences so far.
Blocks without presences are then assigned to the fold with the least pseudo-absences so far, which balances the land area tested per fold.
Each probe is trained on presences and pseudo-absences from four folds and evaluated on presences and pseudo-absences on the remaining fold.
Metrics are averaged over folds and then over species.

\paragraph{Few-shot runs.}
The training presences of each fold are limited to $N$ = 1, 2, 3, 5, 8, 10, 15, 20, 30, 40, 50, 75, or 100.
We perform three runs per species, drawing 100 random presence locations for each.
Within a run, the selected presences for smaller $N$ are subsets of the presences used for larger $N$.
Training pseudo-absence number and removal are recomputed from the $N$ training presences using the process described in ``Pseudo-absences in the UK and California'', above.
The test set presences and pseudo-negatives are selected using all available data and are identical for every $N$ and draw.
We average across runs and then species.
We evaluate only those species that have at least 100 training presences in total, which ends up with 34 of the 37 UK species being used for few-shot experiments. 

\paragraph{Global dataset.}
For the global dataset there is no cross-validation, since the evaluation labels (IUCN range maps) are independent of the training data (iNaturalist observations).
The training presences are the occupied H3 resolution five cells described in Section~\ref{app:datasets}.
Pseudo-absences are $\mathrm{clip}(5n,1000,10000)$ cells drawn uniformly from the 525{,}010 cells which are valid for all feature sets.
All of the valid cells are used during evaluation, with cells whose centre lies inside the range map counted as presences and all others as absences.
AP and AUC are computed per species over all cells and averaged over the 3{,}052 species.

\paragraph{Metrics.}
Average Precision (\emph{AP}) is the average precision of the ranking of test locations, and area under ROC curve (\emph{AUC}) the probability that a random presence outscores a random absence.
\emph{Distance-matched AUC} follows the pairwise distance sampling of~\cite{hijmans2012cross}: it pairs each test presence with the test absence whose distance to the nearest training presence is most similar, and only evaluates those test absences that are paired with a presence.
We apply the same pairing to AP (\emph{distance-matched AP}); see Appendix~\ref{app:results}.
Compared to typical AUC and AP, these metrics reward differentiating between presences and absences close to the training data, and cares less about performance on ``easy'' absences that are very far from the training data.
Features based on geographic location tend to perform well on these easy absences, while features based on environmental covariates are often better at differentiating presences and absences near the training data, while performing worse on ``easy'' absences as distant locations may have similar environmental features, but are very distinct using location features.
Which metric is preferable may depend on the specific use-case being considered.

\section{Additional results}
\label{app:results} %

\begin{figure}[h]
\centering
\includegraphics[width=1.0\textwidth]{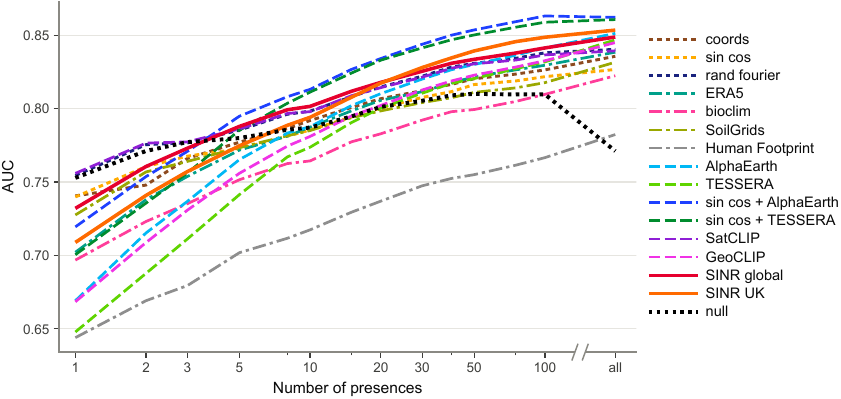}
\caption{Few-shot evaluation for the UK where we vary the amount of presence observations available during training. 
AUC results are reported using a Random Forest probe, averaged over the 34 species with enough presences. `all' uses every training presence.}
\label{fig:uk_few_shot_auc} %
\end{figure}

\begin{figure}[h]
\centering
\includegraphics[width=1.0\textwidth]{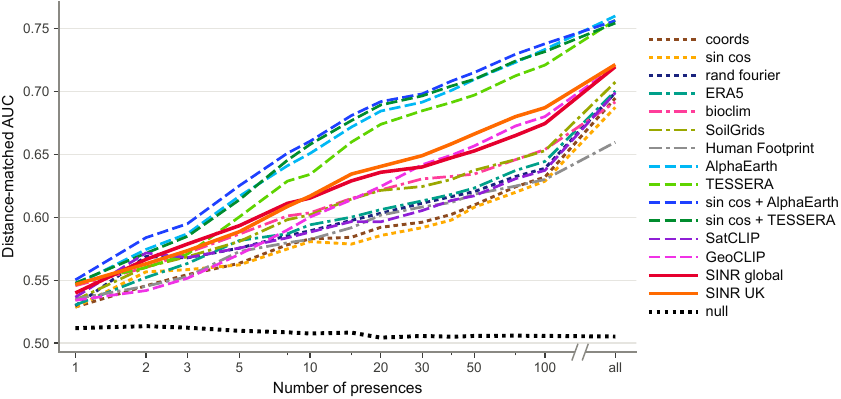}
\caption{Few-shot evaluation for the UK where we vary the amount of presence observations available during training. 
Distance-matched AUC results are reported using a Random Forest probe, averaged over the 34 species with enough presences. `all' uses every training presence.}
\label{fig:uk_few_shot_dm_auc} %
\end{figure}

\begin{figure}[h]
\centering
\includegraphics[width=1.0\textwidth]{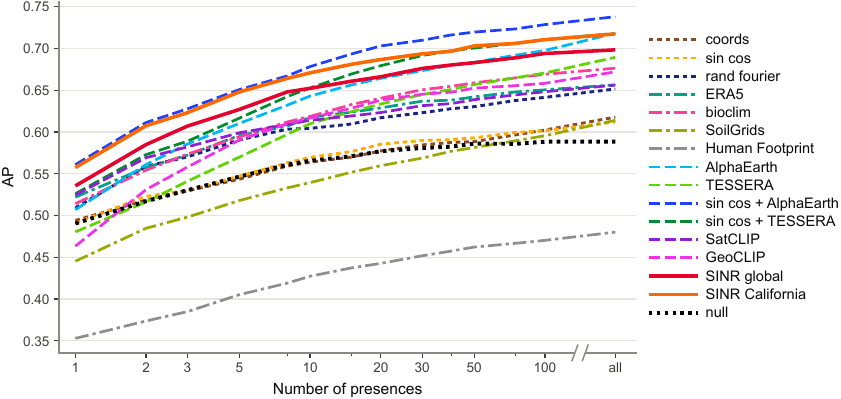}
\caption{Few-shot evaluation for California where we vary the amount of presence observations available during training.
AP results are reported using a Random Forest probe, averaged over the 40 species with enough presences. `all' uses every training presence.}
\label{fig:ca_few_shot}
\end{figure}

\begin{figure}[h]
\centering
\includegraphics[width=1.0\textwidth]{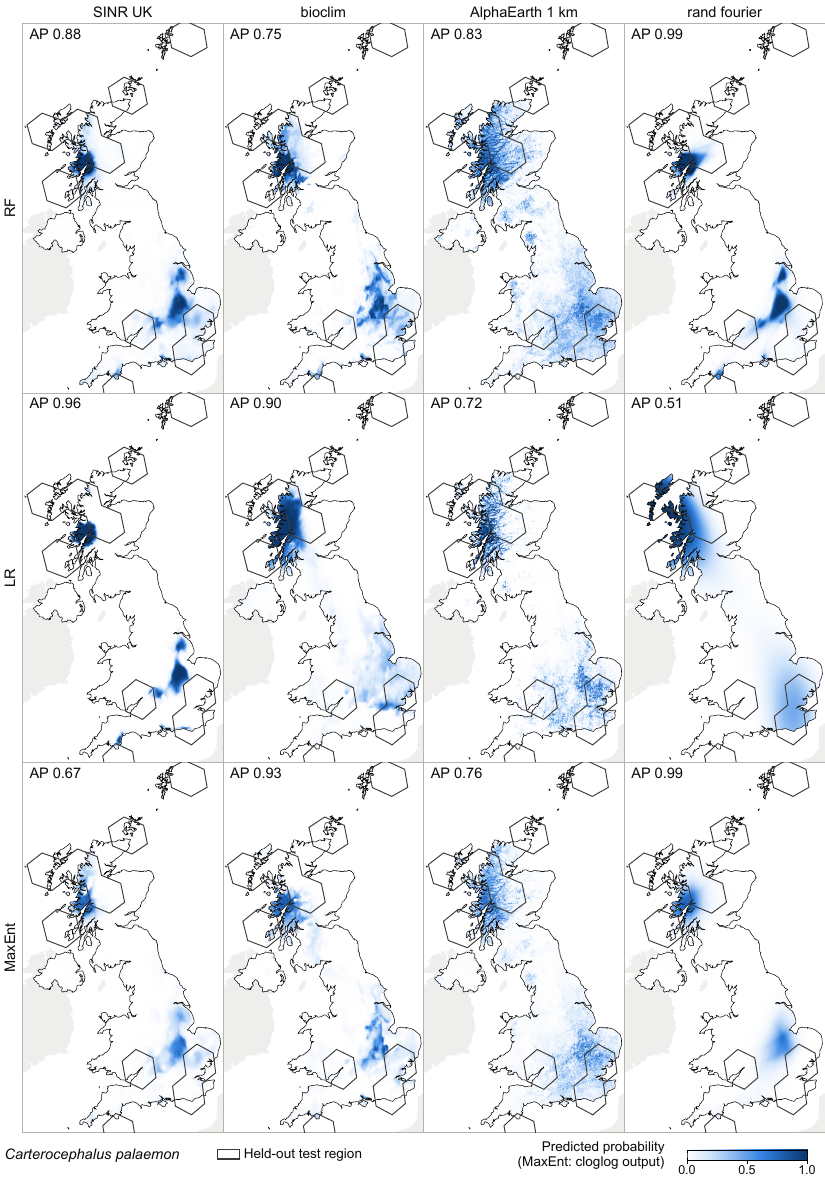}
\caption{Predicted ranges for the butterfly \emph{Carterocephalus palaemon} in the UK with RF (top), LR (middle), and MaxEnt (bottom) probes on four feature sets (SINR UK, bioclim, AlphaEarth 1\,km and random fourier features). Each map comes from a probe of one cross-validation split. The outlined hexagons mark its held-out test region and AP is the score on that region.}
\label{fig:uk_qual_2} %
\end{figure}

\begin{figure}[h]
\centering
\includegraphics[width=1.0\textwidth]{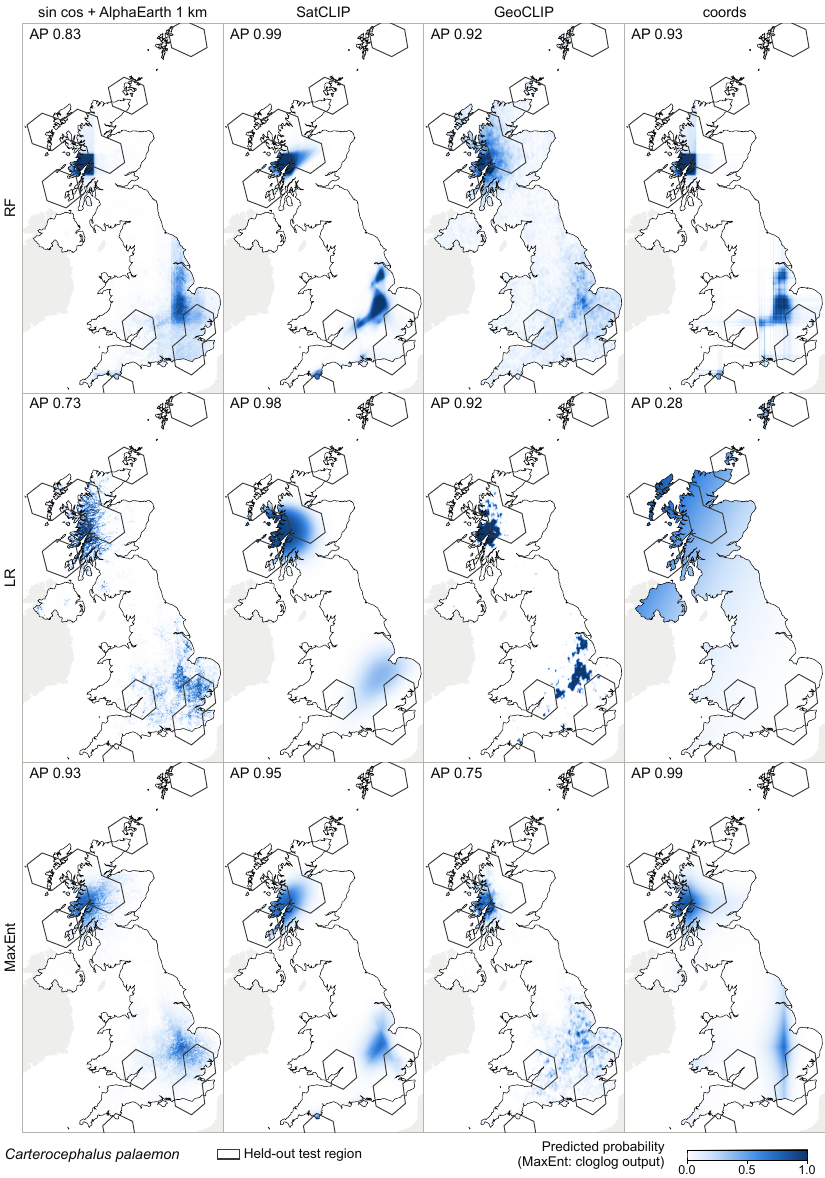}
\caption{Predicted ranges for the butterfly \emph{Carterocephalus palaemon} in the UK with RF (top), LR (middle), and MaxEnt (bottom) probes on four feature sets (sin cos + AlphaEarth 1\,km, SatCLIP, GeoCLIP and coords). Each map comes from a probe of one cross-validation split. The outlined hexagons mark its held-out test region and AP is the score on that region.}
\label{fig:uk_qual_3} %
\end{figure}

\begin{figure}[h]
\centering
\includegraphics[width=1.0\textwidth]{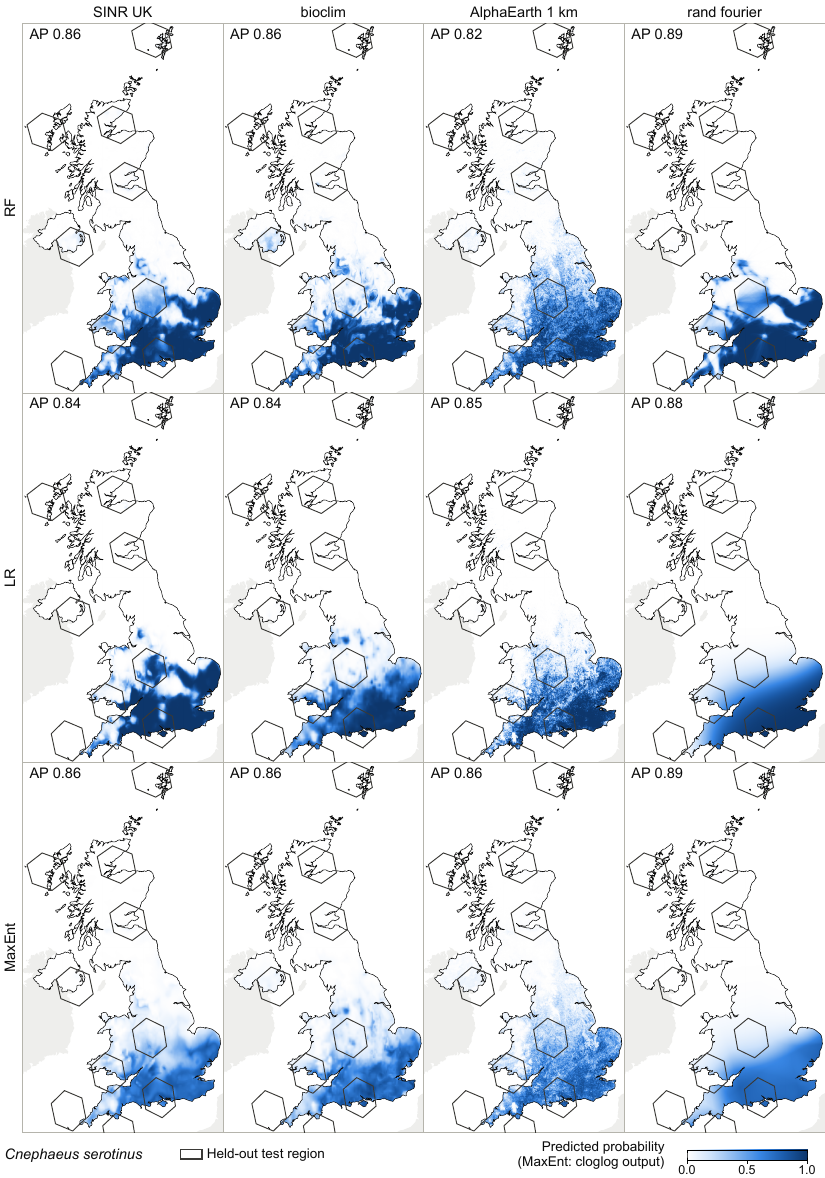}
\caption{Predicted ranges for the bat \emph{Cnephaeus serotinus} in the UK with RF (top), LR (middle), and MaxEnt (bottom) probes on four feature sets (SINR UK, bioclim, AlphaEarth 1\,km and rand fourier). Each map comes from a probe of one cross-validation split. The outlined hexagons mark its held-out test region and AP is the score on that region.}
\label{fig:uk_qual_4} %
\end{figure}

\begin{figure}[h]
\centering
\includegraphics[width=1.0\textwidth]{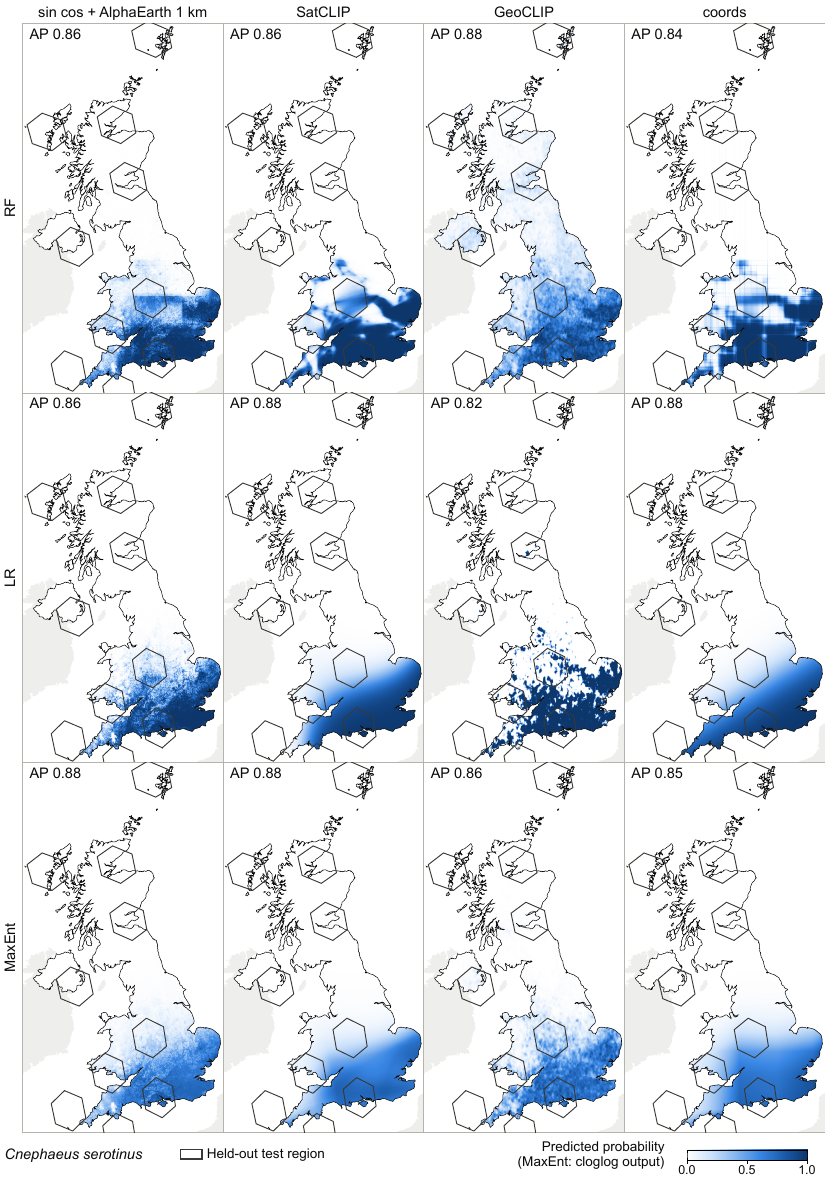}
\caption{Predicted ranges for the bat \emph{Cnephaeus serotinus} in the UK with RF (top), LR (middle), and MaxEnt (bottom) probes on four feature sets (sin cos + AlphaEarth 1\,km, SatCLIP, GeoCLIP and coords). Each map comes from a probe of one cross-validation split. The outlined hexagons mark its held-out test region and AP is the score on that region.}
\label{fig:uk_qual_5} %
\end{figure}

\begin{figure}[h]
\centering
\includegraphics[width=1.0\textwidth]{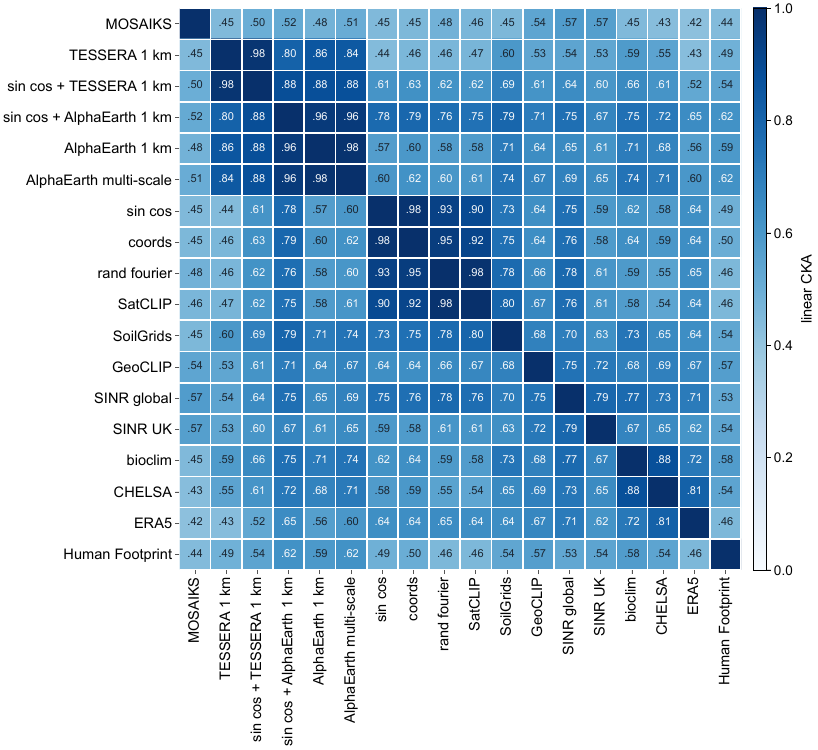}
\caption{Linear CKA~\cite{kornblith2019similarity} between the feature sets in the UK, computed at 13,437 grid points. Values close to one  indicate that two feature sets have very similar linear structure. Rows and columns are ordered by hierarchical clustering.}
\label{fig:uk_cka} %
\end{figure}

\begin{figure}[h]
\centering
\includegraphics[width=1.0\textwidth]{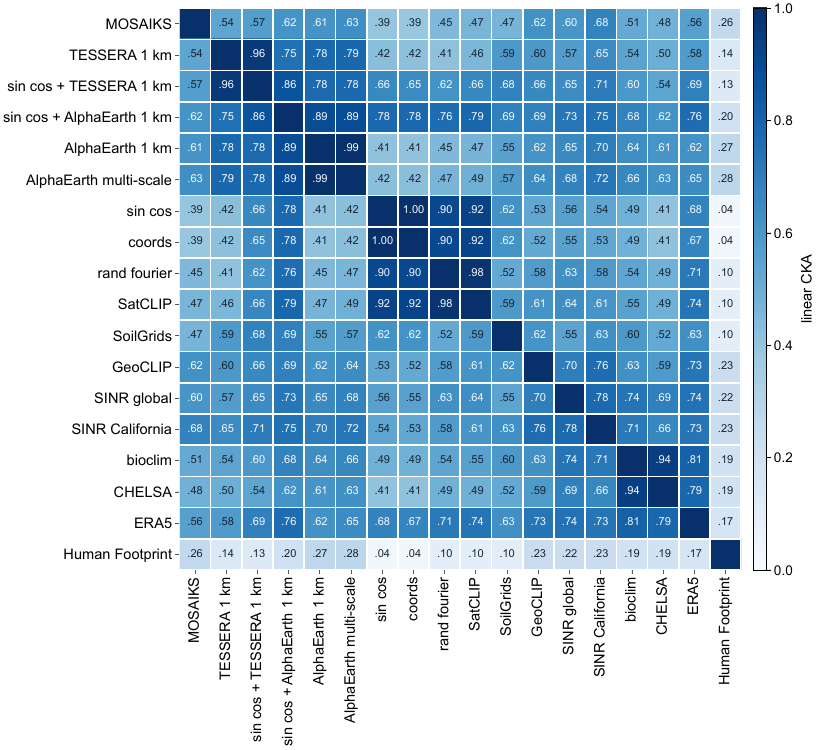}
\caption{Linear CKA~\cite{kornblith2019similarity} between the feature sets in California, computed at 16,584 grid points. Values close to one  indicate that two feature sets have very similar linear structure. Rows and columns follow the UK CKA order.}
\label{fig:ca_cka} %
\end{figure}

\begin{figure}[h]
\centering
\includegraphics[width=1.0\textwidth]{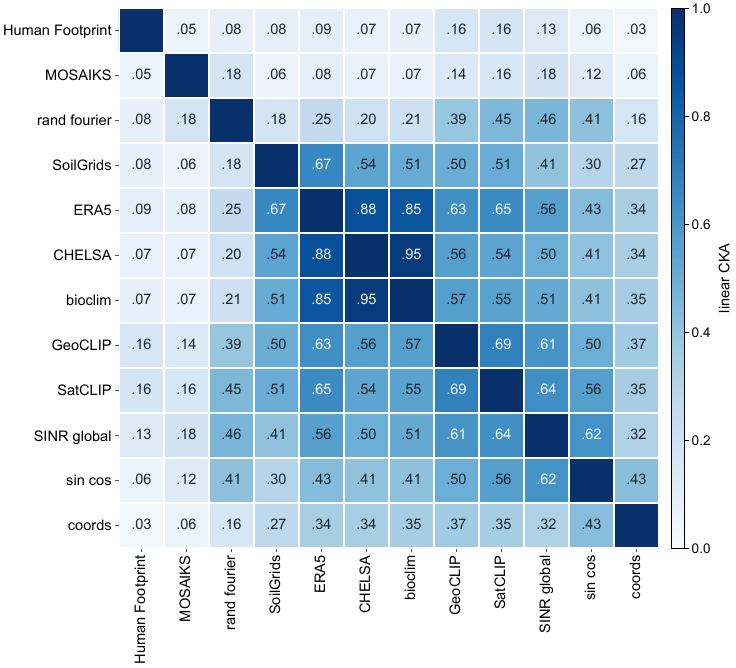}
\caption{Linear CKA~\cite{kornblith2019similarity} between the feature sets at the global scale, computed at 20,000 randomly chosen evaluation cells. Values close to one  indicate that two feature sets have very similar linear structure; rows and columns are ordered by hierarchical clustering.}
\label{fig:gl_cka} %
\end{figure}

\begin{figure}[h]
\centering
\includegraphics[width=1.0\textwidth]{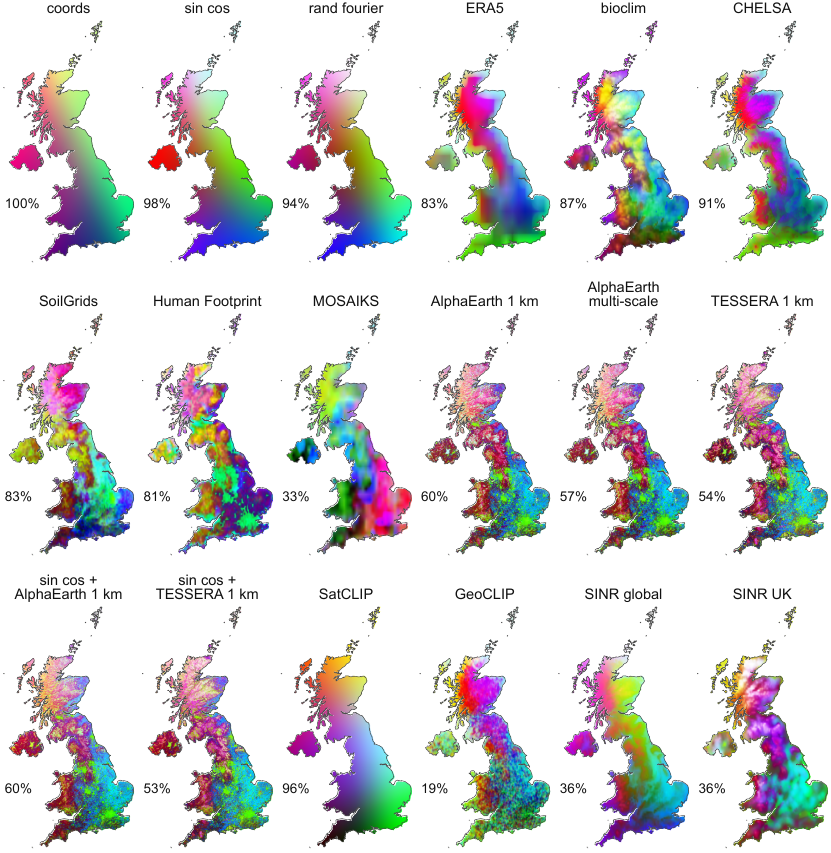}
\caption{The first three principal components of each feature set over the UK, shown as red, green, and blue. Similar colours within a panel indicate similar values on these components. The percentage shown is the variance explained by the three components.}
\label{fig:uk_pca} %
\end{figure}

\begin{figure}[h]
\centering
\includegraphics[width=1.0\textwidth]{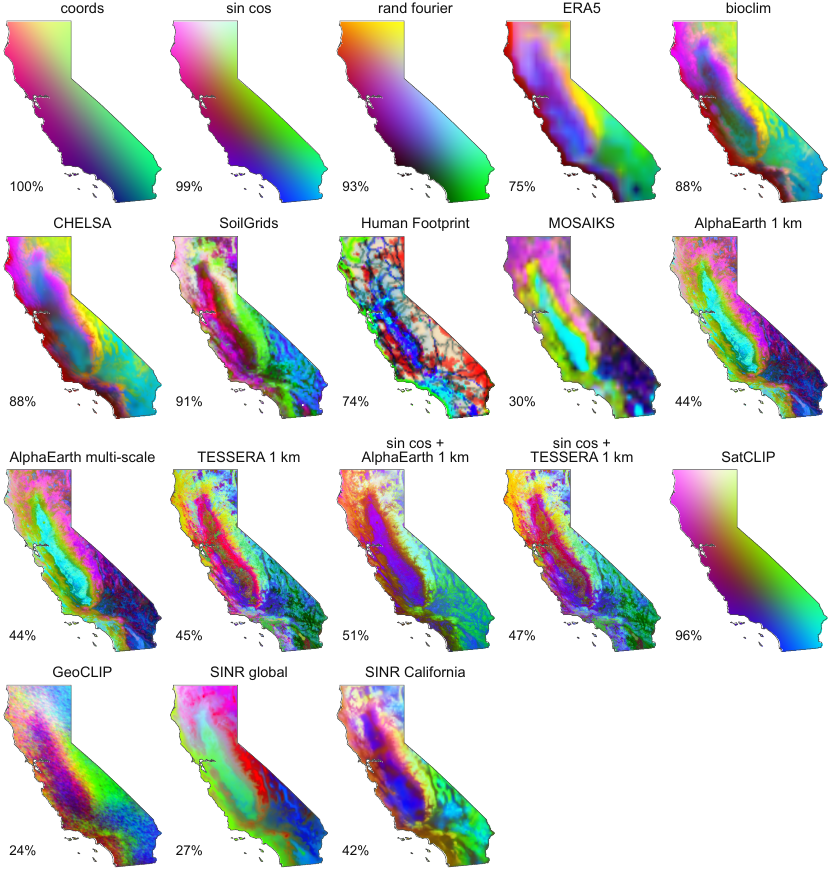}
\caption{The first three principal components of each feature set over California, shown as red, green, and blue. Similar colours within a panel indicate similar values on these components. The percentage shown is the variance explained by the three components.}
\label{fig:ca_pca} %
\end{figure}

\begin{figure}[h]
\centering
\includegraphics[width=1.0\textwidth]{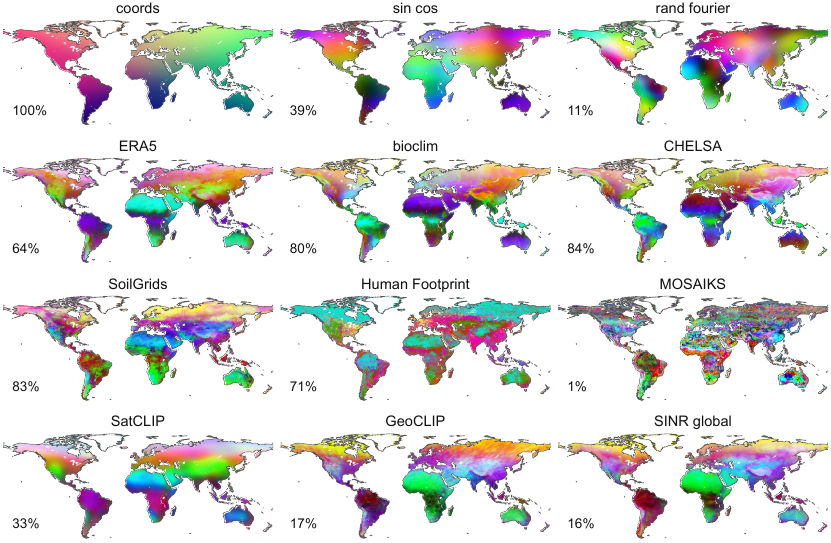}
\caption{The first three principal components of each feature set over the global evaluation cells, shown as red, green, and blue. Similar colours within a panel indicate similar values on these components. The percentage shown is the variance explained by the three components.}
\label{fig:global_pca} %
\end{figure}

This appendix complements Table~\ref{tab:main_res} with results using additional metrics (Tables~\ref{tab:app_res_auc}--\ref{tab:app_res_dmap}), few-shot results for AUC and distance-matched AUC and for California (Figures~\ref{fig:uk_few_shot_auc}--\ref{fig:ca_few_shot}), additional qualitative examples (Figures~\ref{fig:uk_qual_2}--\ref{fig:uk_qual_5}), and a comparison of the feature sets themselves (Figures~\ref{fig:uk_cka}--\ref{fig:global_pca}).
The tables follow the layout of Table~\ref{tab:main_res}: the best result in each column is in bold, ``--'' marks combinations that were not run, and $^{\dagger}$ marks MaxEnt results for which one to three species could not be fit.
The datasets, features, probes and metrics are described in Appendix~\ref{app:methods}.

\paragraph{AUC.}
Table~\ref{tab:app_res_auc} reports the area under the ROC curve.
Globally, AUC is close to its maximum for almost every feature set (0.93--0.98 with RF, except Human Footprint at 0.81), and no feature set exceeds the null baseline (0.99).
This is because most of the 525,010 evaluation cells are far from any training observation of a given species and are easy to reject on distance alone.
In the UK and California, where test absences are restricted to locations at which other species of the same group were recorded, the null baseline is lower (0.78 and 0.85) and all but six of the 108 feature--probe pairs exceed it.

\paragraph{Distance-matched metrics.}
While AUC and AP are frequently used, a model can perform well on these metrics simply by predicting presence close to the training observations, which is exactly what the null baseline does.
This may be more impactful for global-scale evaluations (e.g., using IUCN range maps~\cite{cole2023spatial} or iNaturalist observations~\cite{cole2023spatial,klemmer2025satclip}) which are common in global-change and macroecological studies~\cite{elithreview} and machine-learning based approaches to species distribution modelling.
Regional evaluations, widely used in studies focused on ecological understanding or conservation planning~\cite{elithreview} and in traditional statistical approaches, may be less affected by this.
To measure what a feature set adds beyond proximity, we follow the pairwise distance sampling of~\cite{hijmans2012cross}, which was proposed to remove this ``spatial sorting bias'': every test presence is paired with the test absence whose distance to the nearest training presence is most similar, and the metric is computed on these pairs only.
We report the resulting ``distance-matched'' AUC (dAUC, Table~\ref{tab:app_res_dmauc}) and, as our own extension, a distance-matched AP (dAP, Table~\ref{tab:app_res_dmap}).
By construction, the null baseline is close to chance on dAUC (0.50--0.52).
On dAP it is close to, but not exactly at, chance (0.50--0.56).

The distance-matched metrics change the ranking of the feature sets.
On dAUC every feature set exceeds the null baseline in all three datasets.
Globally, the climate rasters are best (ERA5, bioclim and CHELSA reach 0.68 dAUC with RF, compared to 0.60 for raw coordinates), although they are among the weakest feature sets under AP.
In the UK and California the satellite embeddings lead (AlphaEarth and TESSERA, with or without sin cos: 0.72--0.76 dAUC in the UK and 0.76--0.82 in California).
In California the ordering below them is also clear: the SINR models come next (0.71--0.73), followed by bioclim and CHELSA (0.69--0.71), while the feature sets that mostly encode location (coords, sin cos, rand fourier and SatCLIP) are among the lowest (0.64--0.67).
In the UK the remaining feature sets lie close together (0.65--0.72), without a consistent ordering across probes.
In other words, location-based encodings mainly help to reject absences that are far from the known range, whereas satellite features, and globally the climate features, are better at separating presences from nearby absences.
Which behaviour is preferable depends on the application.

\paragraph{Few-shot results.}
Figures~\ref{fig:uk_few_shot_auc} and~\ref{fig:uk_few_shot_dm_auc} repeat the few-shot experiment of Figure~\ref{fig:uk_few_shot} for AUC and dAUC.
All three UK figures average over the 34 species with enough presences, so their values at `all' (every training presence) are higher than the 37-species means in the tables (e.g., 0.76 compared to 0.70 AP for SINR UK).
Under AUC the picture is similar to AP: with three or fewer training presences at most two of the 15 feature sets exceed the null baseline, and all of them do once every presence is used (all but Human Footprint under AP).
Location-based features (coords, sin cos and random fourier features) perform comparatively well at low number of presences, along with SatCLIP, which at the regional scale resembles these features (see Figures~\ref{fig:uk_cka}, \ref{fig:ca_cka}, \ref{fig:uk_pca}, \ref{fig:ca_pca}).

Under dAUC the null baseline stays at chance for every number of presences, and all feature sets exceed it from a single presence onwards.
The ordering also changes: the AlphaEarth and TESSERA features are ahead of the SINR models from about five presences (e.g., 0.73 for AlphaEarth 1\,km compared to 0.69 for SINR UK at 100 presences), whereas under AP SINR UK is ahead of AlphaEarth 1\,km at every number of presences.
Overall, earth embedding features (TESSERA and AlphaEarth) perform very well under this metric, and the concatenation with location-based features provides less benefit.
location-based features alone perform much worse than for AUC, even with few presences, as their ability to identify `easy' absences far from the training data is not evaluated.

Figure~\ref{fig:ca_few_shot} shows few-shot results for California under AP, averaged over the 40 species with at least 100 train presences.
The null baseline is weaker in California: 11 of the 15 feature sets exceed it from a single presence, and all but Human Footprint do from 75 presences.
Similarly, location-based features also perform worse compared to the UK.
SINR California is ahead of SINR global at every number of presences (by about 0.02 AP), and sin cos + AlphaEarth 1\,km is the best feature set throughout.

\paragraph{Qualitative examples.}
Figures~\ref{fig:uk_qual_2}--\ref{fig:uk_qual_5} show predicted ranges for the two species of Figure~\ref{fig:uk_qual_1}, with one row per probe (RF, LR and MaxEnt) and one column per feature set.
Figures~\ref{fig:uk_qual_2} and~\ref{fig:uk_qual_4} use the feature sets of Figure~\ref{fig:uk_qual_1}; Figures~\ref{fig:uk_qual_3} and~\ref{fig:uk_qual_5} show sin cos + AlphaEarth 1\,km, SatCLIP, GeoCLIP and coords.
Each map is produced by a probe from one cross-validation split (trained on four folds and tested on the fifth): the training and test presences of that split are shown in Figure~\ref{fig:uk_qual_1}, the outlined hexagons mark its held-out test region, and the AP printed on each map is the score on that region.
We see that probes using location-based features (random fourier features, coordinates, SatCLIP) can effectively localize their presence predictions to areas near the training data, while probes using environmental features (bioclim) and and earth embeddings (AlphaEarth 1km) can struggle with this but generate higher frequency predictions.
This leads to location-based features performing comparatively better under AUC and AP, compared to the the distance-matched versions of these metrics, where performance far from training presences is not evaluated. 
Inversely, environmental features, which allow higher frequency changes between presence and absence close to the training data, perform better under the distance-matched metrics.

The choice of probe has a visible effect on the predictions.
Random forests on features that contain coordinates or sin cos transforms of the coordinates produce axis-aligned blocks.
Despite these visible artifacts, these approaches can perform well, with a random forest probe using sin cos + AlphaEarth 1km features outperforming all other combinations of features and probes in Table~\ref{tab:main_res}.
We also see that logistic regression on raw coordinates can only express a single gradient.

\paragraph{Similarity between feature sets.}
Figures~\ref{fig:uk_cka}--\ref{fig:gl_cka} report the linear centered kernel alignment (CKA)~\cite{kornblith2019similarity} between each pair of feature sets, computed on features standardized per dimension at 13,437 grid points in the UK, 16,584 grid points in California, and 20,000 randomly chosen evaluation cells globally.
A value of 1 means that two feature sets are identical up to rotation and scaling, and 0 that they are linearly unrelated.
A high value indicates shared linear structure, and not necessarily that two feature sets are equally useful for species distribution modelling.

At the regional scale, coordinate encodings (coords, sin cos, random fourier features) and SatCLIP are highly similar (CKA 0.90--1.00), however, globally, these features are more distinct.
This can also be seen in the PCA visualizations in Figures~\ref{fig:uk_pca}--\ref{fig:global_pca}.
AlphaEarth and TESSERA are similar to each other (0.86 and 0.78) but less similar to the climate rasters (ERA5, CHELSA, bioclim) (0.43--0.71), which are similar to each other.
Despite the SINR UK and SINR California models having the same architecture and training setup as SINR global, and using regional subsets of the same training data, they are not as similar as the earth embeddings which come from distinct models and training data.
The most distinct feature set is MOSAIKS in the UK (at most 0.57 with any other feature set) and Human Footprint in California (at most 0.28).

Globally, the similarities are lower overall: the three climate features are still close (0.85--0.95), but satCLIP and the coordinate encodings are not as similar to each other (e.g., 0.43 between coords and sin cos).
Human Footprint and MOSAIKS remain very dissimilar to all other features.

\paragraph{Visualising the feature spaces.}
Figures~\ref{fig:uk_pca}--\ref{fig:global_pca} visualize each feature set by projecting it onto its first three principal components (computed over all mapped locations after standardizing each dimension), rank-transforming each component to $[0,1]$, and displaying the three components as the red, green, and blue channels.
Locations with similar colors therefore have similar values on these three components, but colors are not comparable between panels.
The percentage in each panel is the share of the variance of that feature set explained by the three components, however coords has only two dimensions, so its third channel is constant.
The maps use a 0.025$^\circ$ grid in the UK (53,677 points) and California (66,514 points), and all 525,010 evaluation cells globally.
We can see some features are relatively smooth, including coordinates, sin cos, random fourier features and SatCLIP, while environment features and earth embeddings show sharper features.
This is most evident on the smaller regional figures. 
The smooth or low dimensional feature sets are largely described by three components within a region (e.g., 96\% for SatCLIP and 87\% for bioclim in the UK) while the relatively high dimensional learned embeddings spread their variance over many more directions (e.g., 60\% for AlphaEarth 1\,km, 36\% for SINR UK and 19\% for GeoCLIP in the UK).
AlphaEarth, TESSERA and MOSAIKS, which use satellite imagery, pick up heavily urbanized areas in the UK, which are not evident from environment embeddings or the SINR models which are trained only on species observations.
SatCLIP is also trained on satellite imagery, but this training data is relatively sparse and is not used during inference and so SatCLIP does not seem to capture this relatively high frequency information.

\begin{table}[h]
  \caption{Quantitative evaluation of different location encodings using Random Forest (RF), Logistic Regression (LR), or MaxEnt (ME) probes on the task of species range estimation. Performance is reported using area under ROC curve (AUC).}
  \label{tab:app_res_auc}
\resizebox{\textwidth}{!}{
\begin{tabular}{lcr|rrl|rrl|rrl}
 & &  & \multicolumn{3}{|c|}{\textbf{Global}} & \multicolumn{3}{|c|}{\textbf{UK}} & \multicolumn{3}{|c}{\textbf{California}} \\
\textbf{Features} & \textbf{Type} & \textbf{Dim} & \textbf{RF} & \textbf{LR} & \textbf{ME} & \textbf{RF} & \textbf{LR} & \textbf{ME} & \textbf{RF} & \textbf{LR} & \textbf{ME} \\ \hline
coords & loc & 2 & \textbf{0.98} & 0.89 & 0.98$^{\dagger}$ & 0.83 & 0.80 & 0.83 & 0.87 & 0.83 & 0.87 \\
sin cos & loc & 16 & \textbf{0.98} & \textbf{0.98} & \textbf{0.99} & 0.82 & 0.82 & 0.82 & 0.87 & 0.86 & 0.87 \\
rand fourier & loc & 128 & 0.97 & 0.96 & 0.97 & 0.84 & 0.83 & 0.84 & 0.89 & 0.88 & 0.87 \\
ERA5 & env & 117 & 0.96 & 0.96 & 0.95 & 0.83 & 0.84 & 0.81 & 0.90 & 0.88 & 0.87 \\
bioclim & env & 20 & 0.96 & 0.93 & 0.96 & 0.82 & 0.82 & 0.82 & 0.90 & 0.89 & 0.91 \\
CHELSA & env & 19 & 0.95 & 0.92 & 0.95 & 0.81 & 0.81 & 0.82 & 0.89 & 0.87 & 0.89 \\
SoilGrids & env & 14 & 0.93 & 0.90 & 0.93$^{\dagger}$ & 0.83 & 0.80 & 0.81 & 0.87 & 0.84 & 0.87 \\
Human Footprint & env & 9 & 0.81 & 0.78 & 0.81$^{\dagger}$ & 0.78 & 0.77 & 0.78 & 0.74 & 0.72 & 0.74$^{\dagger}$ \\
MOSAIKS & ee & 512 & 0.93 & 0.91 & 0.88$^{\dagger}$ & 0.84 & 0.80 & 0.81 & 0.89 & 0.86 & 0.89 \\
AlphaEarth 1 km & ee & 64 & -- & -- & -- & 0.84 & 0.85 & 0.83 & 0.92 & 0.92 & 0.92 \\
AlphaEarth multi-scale & ee & 192 & -- & -- & -- & 0.85 & 0.85 & 0.83 & \textbf{0.93} & \textbf{0.93} & \textbf{0.93} \\
TESSERA 1 km & ee & 128 & -- & -- & -- & 0.84 & 0.85 & 0.83 & 0.91 & 0.91 & 0.90 \\
sin cos + AlphaEarth 1 km & loc + ee & 80 & -- & -- & -- & \textbf{0.86} & \textbf{0.86} & \textbf{0.86} & \textbf{0.93} & \textbf{0.93} & \textbf{0.93} \\
sin cos + TESSERA 1 km & loc + ee & 144 & -- & -- & -- & \textbf{0.86} & \textbf{0.86} & 0.83 & 0.92 & 0.92 & 0.90 \\
SatCLIP & ee & 256 & \textbf{0.98} & \textbf{0.98} & 0.98 & 0.84 & 0.84 & 0.83 & 0.89 & 0.88 & 0.87 \\
GeoCLIP & ee & 512 & \textbf{0.98} & 0.97 & 0.98 & 0.85 & 0.82 & 0.84 & 0.90 & 0.87 & 0.89 \\
SINR global & sdm & 256 & \textbf{0.98} & 0.96 & 0.98 & 0.85 & 0.82 & 0.83 & 0.91 & 0.89 & 0.90 \\
SINR UK / California & sdm & 256 & -- & -- & -- & 0.85 & 0.83 & 0.84 & 0.92 & 0.91 & 0.91 \\
\hline
null (dist. to nearest obs.) & baseline & -- & \multicolumn{3}{|c|}{0.99} & \multicolumn{3}{|c|}{0.78} & \multicolumn{3}{|c}{0.85} \\
\end{tabular}
}
\end{table}

\begin{table}[h]
  \caption{Quantitative evaluation of different location encodings using Random Forest (RF), Logistic Regression (LR), or MaxEnt (ME) probes on the task of species range estimation. Performance is reported using ``distance-matched'' area under ROC curve (dAUC, i.e., AUC computed after pairing every test presence with the test absence at the most similar distance to the training presences, following~\cite{hijmans2012cross}.}
  \label{tab:app_res_dmauc}
\resizebox{\textwidth}{!}{
\begin{tabular}{lcr|rrl|rrl|rrl}
 & &  & \multicolumn{3}{|c|}{\textbf{Global}} & \multicolumn{3}{|c|}{\textbf{UK}} & \multicolumn{3}{|c}{\textbf{California}} \\
\textbf{Features} & \textbf{Type} & \textbf{Dim} & \textbf{RF} & \textbf{LR} & \textbf{ME} & \textbf{RF} & \textbf{LR} & \textbf{ME} & \textbf{RF} & \textbf{LR} & \textbf{ME} \\ \hline
coords & loc & 2 & 0.60 & 0.51 & 0.60$^{\dagger}$ & 0.68 & 0.66 & 0.69 & 0.64 & 0.64 & 0.65 \\
sin cos & loc & 16 & 0.62 & 0.61 & 0.61 & 0.68 & 0.69 & 0.68 & 0.64 & 0.65 & 0.65 \\
rand fourier & loc & 128 & 0.63 & 0.59 & 0.63 & 0.68 & 0.70 & 0.68 & 0.67 & 0.66 & 0.65 \\
ERA5 & env & 117 & \textbf{0.68} & \textbf{0.64} & 0.65 & 0.69 & 0.72 & 0.67 & 0.68 & 0.67 & 0.63 \\
bioclim & env & 20 & \textbf{0.68} & 0.62 & \textbf{0.66} & 0.68 & 0.70 & 0.69 & 0.71 & 0.70 & 0.71 \\
CHELSA & env & 19 & \textbf{0.68} & 0.60 & \textbf{0.66} & 0.68 & 0.68 & 0.68 & 0.71 & 0.69 & 0.70 \\
SoilGrids & env & 14 & 0.65 & 0.62 & 0.64$^{\dagger}$ & 0.69 & 0.66 & 0.67 & 0.70 & 0.66 & 0.68 \\
Human Footprint & env & 9 & 0.55 & 0.53 & 0.55$^{\dagger}$ & 0.65 & 0.65 & 0.66 & 0.64 & 0.61 & 0.62$^{\dagger}$ \\
MOSAIKS & ee & 512 & 0.65 & 0.61 & 0.62$^{\dagger}$ & 0.70 & 0.68 & 0.68 & 0.67 & 0.65 & 0.67 \\
AlphaEarth 1 km & ee & 64 & -- & -- & -- & 0.74 & 0.75 & 0.72 & \textbf{0.80} & 0.80 & 0.79 \\
AlphaEarth multi-scale & ee & 192 & -- & -- & -- & 0.74 & \textbf{0.76} & 0.73 & \textbf{0.80} & \textbf{0.82} & \textbf{0.81} \\
TESSERA 1 km & ee & 128 & -- & -- & -- & 0.74 & 0.75 & 0.72 & 0.76 & 0.76 & 0.76 \\
sin cos + AlphaEarth 1 km & loc + ee & 80 & -- & -- & -- & \textbf{0.75} & \textbf{0.76} & \textbf{0.75} & 0.78 & 0.80 & 0.78 \\
sin cos + TESSERA 1 km & loc + ee & 144 & -- & -- & -- & \textbf{0.75} & \textbf{0.76} & 0.72 & 0.76 & 0.77 & 0.76 \\
SatCLIP & ee & 256 & 0.65 & 0.60 & 0.63 & 0.68 & 0.70 & 0.70 & 0.67 & 0.66 & 0.65 \\
GeoCLIP & ee & 512 & 0.64 & 0.60 & 0.63 & 0.72 & 0.68 & 0.72 & 0.71 & 0.66 & 0.71 \\
SINR global & sdm & 256 & 0.67 & 0.61 & \textbf{0.66} & 0.70 & 0.69 & 0.68 & 0.72 & 0.71 & 0.71 \\
SINR UK / California & sdm & 256 & -- & -- & -- & 0.71 & 0.70 & 0.71 & 0.73 & 0.73 & 0.72 \\
\hline
null (dist. to nearest obs.) & baseline & -- & \multicolumn{3}{|c|}{0.50} & \multicolumn{3}{|c|}{0.51} & \multicolumn{3}{|c}{0.52} \\
\end{tabular}
}
\end{table}

\begin{table}[]
  \caption{Quantitative evaluation of different location encodings using Random Forest (RF), Logistic Regression (LR), or MaxEnt (ME) probes on the task of species range estimation. Performance is reported using  ``distance-matched'' Average Precision (dAP), computed on the same matched pairs as dAUC.}
  \label{tab:app_res_dmap} %
\resizebox{\textwidth}{!}{
\begin{tabular}{lcr|rrl|rrl|rrl}
 & &  & \multicolumn{3}{|c|}{\textbf{Global}} & \multicolumn{3}{|c|}{\textbf{UK}} & \multicolumn{3}{|c}{\textbf{California}} \\
\textbf{Features} & \textbf{Type} & \textbf{Dim} & \textbf{RF} & \textbf{LR} & \textbf{ME} & \textbf{RF} & \textbf{LR} & \textbf{ME} & \textbf{RF} & \textbf{LR} & \textbf{ME} \\ \hline
coords & loc & 2 & 0.59 & 0.50 & 0.58$^{\dagger}$ & 0.66 & 0.66 & 0.67 & 0.64 & 0.64 & 0.65 \\
sin cos & loc & 16 & 0.62 & 0.60 & 0.60 & 0.65 & 0.68 & 0.67 & 0.65 & 0.65 & 0.66 \\
rand fourier & loc & 128 & 0.64 & 0.58 & 0.63 & 0.66 & 0.69 & 0.67 & 0.67 & 0.66 & 0.65 \\
ERA5 & env & 117 & \textbf{0.68} & \textbf{0.62} & 0.63 & 0.67 & 0.71 & 0.66 & 0.68 & 0.67 & 0.64 \\
bioclim & env & 20 & 0.67 & 0.60 & \textbf{0.65} & 0.67 & 0.69 & 0.68 & 0.71 & 0.70 & 0.71 \\
CHELSA & env & 19 & \textbf{0.68} & 0.58 & 0.64 & 0.67 & 0.68 & 0.67 & 0.70 & 0.69 & 0.69 \\
SoilGrids & env & 14 & 0.65 & 0.61 & 0.62$^{\dagger}$ & 0.68 & 0.66 & 0.66 & 0.70 & 0.66 & 0.68 \\
Human Footprint & env & 9 & 0.56 & 0.53 & 0.55$^{\dagger}$ & 0.64 & 0.65 & 0.65 & 0.67 & 0.65 & 0.66$^{\dagger}$ \\
MOSAIKS & ee & 512 & 0.65 & 0.59 & 0.62$^{\dagger}$ & 0.68 & 0.67 & 0.67 & 0.67 & 0.66 & 0.67 \\
AlphaEarth 1 km & ee & 64 & -- & -- & -- & \textbf{0.73} & 0.75 & 0.72 & 0.80 & 0.80 & 0.79 \\
AlphaEarth multi-scale & ee & 192 & -- & -- & -- & \textbf{0.73} & 0.76 & 0.72 & \textbf{0.81} & \textbf{0.83} & \textbf{0.80} \\
TESSERA 1 km & ee & 128 & -- & -- & -- & \textbf{0.73} & 0.75 & 0.71 & 0.77 & 0.76 & 0.76 \\
sin cos + AlphaEarth 1 km & loc + ee & 80 & -- & -- & -- & \textbf{0.73} & \textbf{0.77} & \textbf{0.73} & 0.78 & 0.81 & 0.78 \\
sin cos + TESSERA 1 km & loc + ee & 144 & -- & -- & -- & \textbf{0.73} & 0.75 & 0.72 & 0.77 & 0.77 & 0.76 \\
SatCLIP & ee & 256 & 0.64 & 0.59 & 0.62 & 0.66 & 0.69 & 0.68 & 0.67 & 0.67 & 0.66 \\
GeoCLIP & ee & 512 & 0.64 & 0.60 & 0.62 & 0.70 & 0.67 & 0.70 & 0.71 & 0.67 & 0.71 \\
SINR global & sdm & 256 & 0.67 & 0.60 & \textbf{0.65} & 0.68 & 0.68 & 0.67 & 0.72 & 0.71 & 0.70 \\
SINR UK / California & sdm & 256 & -- & -- & -- & 0.69 & 0.69 & 0.69 & 0.73 & 0.73 & 0.72 \\
\hline
null (dist. to nearest obs.) & baseline & -- & \multicolumn{3}{|c|}{0.50} & \multicolumn{3}{|c|}{0.54} & \multicolumn{3}{|c}{0.56} \\
\end{tabular}
}
\end{table}

\clearpage

\end{document}